%% file: main.tex
\pdfoutput=1

\documentclass[11pt]{article}

\usepackage[preprint]{acl}
\hypersetup{
    pagebackref=true,
    breaklinks=true,
    colorlinks=true,
}
\newcommand{\modelname}{VideoPASTA}

\makeatletter
\renewcommand\paragraph{\@startsection{paragraph}{4}{\z@}%
  {0.1ex \@plus0.2ex \@minus.1ex}
  {-0.2em}
  {\normalfont\normalsize\bfseries}}
\makeatother

\usepackage{tcolorbox}
\usepackage{amsmath}
\usepackage{adjustbox}
\usepackage{times}
\usepackage{amssymb}
\usepackage{multirow}
\usepackage{multicol}
\usepackage{arydshln}
\usepackage{algorithm}   
\usepackage{subcaption}   
\usepackage{algorithm}
\usepackage{algpseudocode}

\usepackage{tabularx}
\usepackage[table]{xcolor}
\usepackage{booktabs}
\usepackage{comment}
\definecolor{iccvblue}{rgb}{0.21,0.49,0.74}
\definecolor{Gray}{gray}{0.5}
\definecolor{LGray}{gray}{0.9}
\definecolor{darkblue}{RGB}{94,110,186}
\definecolor{darkGreen}{RGB}{92, 148, 110}

\definecolor{darkgreen}{RGB}{0,100,0}
\definecolor{lightorange}{RGB}{255,160,0}
\definecolor{darkred}{RGB}{139,0,0}
\definecolor{myblue}{RGB}{14, 121, 178}
\usepackage{colortbl}
\usepackage{pifont}
\usepackage{makecell}
\usepackage{fix-cm} 
\usepackage{newtxmath}

\definecolor{PastaYellow}{RGB}{255,230,190}
\definecolor{GroupColor1}{RGB}{220,230,255}
\definecolor{GroupColor2}{RGB}{220,255,220}
\definecolor{GroupColor3}{RGB}{255,220,200}
\definecolor{GroupColor4}{RGB}{230,220,255}
\definecolor{GroupColor5}{RGB}{210,200,205}

\usepackage{times}
\usepackage{latexsym}
\usepackage[T1]{fontenc}
\usepackage[utf8]{inputenc}
\usepackage{microtype}
\usepackage{inconsolata}
\usepackage{graphicx}

\author{
    Yogesh Kulkarni \qquad 
    Pooyan Fazli \qquad\\
    Arizona State University\\
    {\vspace{0.5em}\normalsize \textcolor{iccvblue}{\url{https://people-robots.github.io/VideoPASTA/}}}
}

\title{\modelname \hspace{-0.2em}~\raisebox{-0.2ex}{\includegraphics[height=1em]{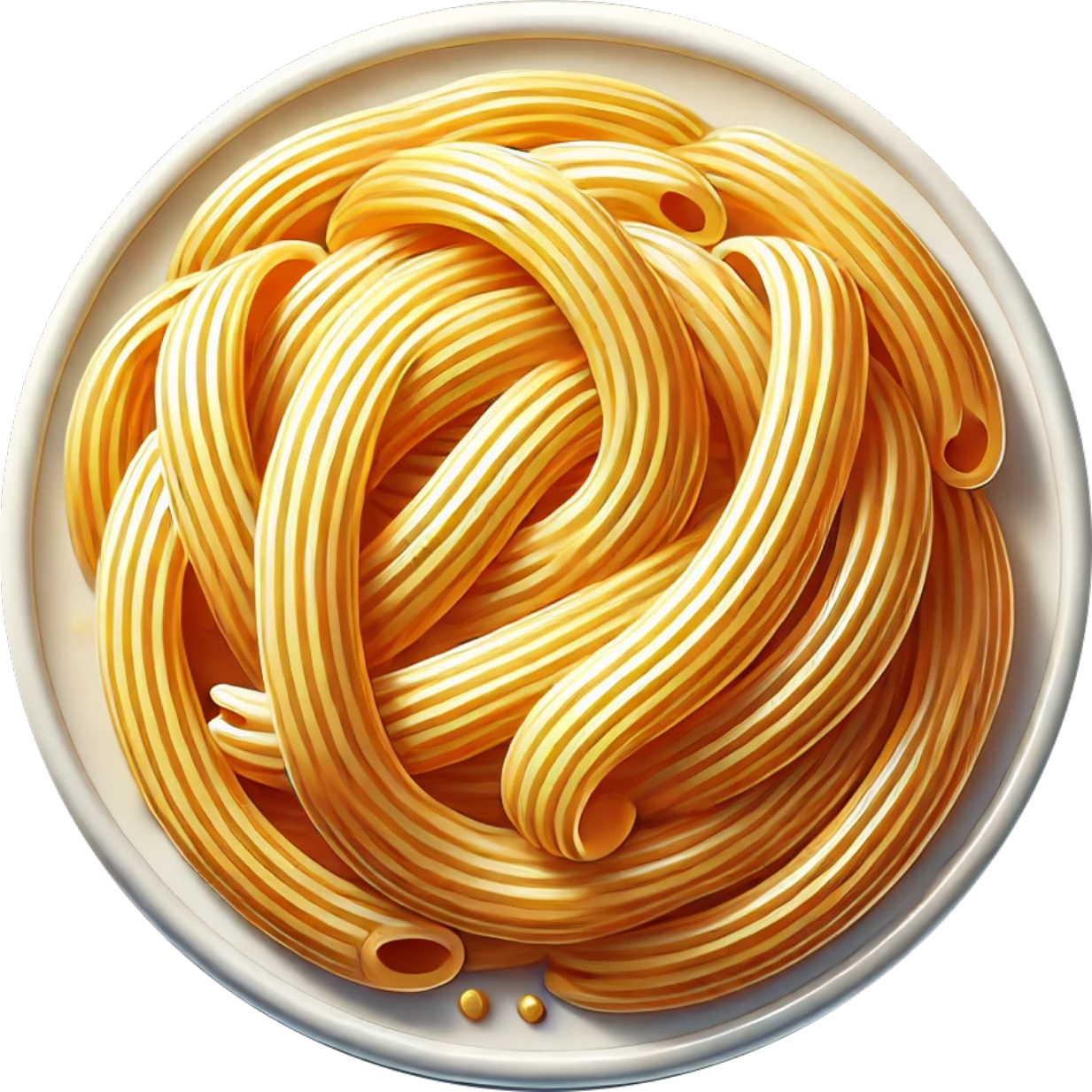}}: 7K Preference Pairs That Matter for Video-LLM Alignment}

\begin{document}
\maketitle
\input{sec/abstract}

\input{sec/intro}

\input{sec/2_related}

\input{sec/4_approach}

\input{sec/5_experiments}

\section*{Limitations}

While \modelname{} demonstrates significant advancements, certain aspects warrant future exploration. The quality and diversity of the generated preference pairs depend on the capabilities of the models used in our pipeline (i.e., query generator, response generator, and verifier). Potential biases or limitations in these foundational models could subtly affect the preference dataset.

\section*{Acknowledgments}
This research was supported by the National Eye Institute (NEI) of the National Institutes of Health (NIH) under award number R01EY034562. The content is solely the responsibility of the authors and does not necessarily represent the official views of the NIH.

\bibliography{custom}
\clearpage
\appendix

\section*{Appendix}
\label{sec:appendix}

\input{sec/X_suppl}

\end{document}

%% file: sec/abstract.tex
\begin{abstract}
Video-language models (Video-LLMs) excel at understanding video content but struggle with spatial relationships, temporal ordering, and cross-frame continuity. To address these limitations, we introduce \textbf{\modelname} \hspace{-2.2mm}~\raisebox{-0.2ex}{\includegraphics[height=1em]{images/pasta.pdf}} (\textbf{P}reference \textbf{A}lignment with \textbf{S}patio-\textbf{T}emporal-Cross Frame \textbf{A}dversaries), a framework that enhances Video-LLMs through targeted preference optimization.\ 
\modelname{} trains models to distinguish accurate video representations from carefully crafted adversarial examples that deliberately violate spatial, temporal, or cross-frame relationships.\ With only 7,020 preference pairs and Direct Preference Optimization, \modelname{} enables models to learn robust representations that capture fine-grained spatial details and long-range temporal dynamics.\ Experiments demonstrate that \modelname{} is model agnostic and significantly improves performance, for example, achieving gains of up to +3.8 percentage points on LongVideoBench, +4.1 on VideoMME, and +4.0 on MVBench, when applied to various state-of-the-art Video-LLMs.\ 
These results demonstrate that targeted alignment, rather than massive pretraining or architectural modifications, effectively addresses core video-language challenges.\ 
Notably, \modelname{} achieves these improvements without any human annotation or captioning, relying solely on 32-frame sampling. This efficiency makes our approach a scalable plug-and-play solution that seamlessly integrates with existing models while preserving their original capabilities.
\end{abstract}

%% file: sec/intro.tex
\input{figures/mvbench_eff}

\section{Introduction}

Recent advances in video language models (Video-LLMs) have enabled efficient video understanding and reasoning, achieving strong performance on tasks like captioning and question answering~\cite{ liLLaVANeXTInterleaveTacklingMultiimage2024, internvideo, internvideo2, qwen2.5-VL, internvl2_5}. However, these models typically rely on large, high-quality annotated datasets and significant computing resources for training. Instruction tuning has emerged as a way to reduce data requirements by fine-tuning models on curated instruction-response pairs~\cite{llava, zhang2024video, videollava, internvl2_5, qwen2vl}. While large instruction datasets have been generated using models like GPT-4V~\cite{zhang2024video}, improvements from training on these datasets remain limited. Video-LLMs still struggle with spatial misalignment, temporal incoherence, and cross-frame disconnections~\cite{choong2024vidhal, hu2025cos, leng2024mitigating, ma2024vista, gunjal2024detecting}. Addressing these issues through human annotation is expensive, as it requires identifying examples with proper grounding and coherence. This suggests that merely scaling models and data alone is insufficient. Instead, the core challenge lies in achieving faithful alignment between model outputs and video content.
Recent work has applied Direct Preference Optimization (DPO) to improve video-language alignment~\cite{ahnISRTAligningLarge2024, zhang2024direct, li2025temporal, rafailovDirectPreferenceOptimization2023}. However, these methods often reinforce existing strengths through collecting more preference data instead of targeting core weaknesses in Video-LLMs. They also rely on proprietary models~\cite{zhang2024direct} or video captioning~\cite{li2025temporal}, limiting scalability.

This raises a key question: \textit{How can we align Video-LLMs to understand spatial, temporal, and cross-frame relationships without human annotations, captions, or proprietary models, while remaining computationally efficient?} To address this, we introduce \textbf{\modelname} \hspace{-0.4em}~\raisebox{-0.2ex}{\includegraphics[height=1em]{images/pasta.pdf}} (\textbf{P}reference \textbf{A}lignment with \textbf{S}patio-\textbf{T}emporal-Cross Frame \textbf{A}dversaries), a model-agnostic framework that improves video-language alignment using targeted preference pairs. \modelname{} contrasts aligned (``preferred'') responses with adversarial ones that capture three common failure modes in video understanding: (1) \textbf{spatial misalignment}, where responses misrepresent object relationships and interactions, (2) \textbf{temporal incoherence}, where responses violate the natural progression of events, and (3) \textbf{cross-frame disconnection}, where responses violate object persistence, character consistency, and narrative progression across more distant parts of a video. By combining DPO with structured preference data, \modelname{} directly tackles these limitations in Video-LLMs. In summary, our contributions are as follows:

\begin{enumerate}
    \item We introduce \modelname{}, a novel, model-agnostic DPO framework that improves video-language alignment by addressing spatial misalignment, temporal incoherence, and cross-frame disconnection, without human annotations, captions, or proprietary models.
    
        \vspace{-0.2cm}
    \item \modelname{} sets a new efficiency benchmark, achieving strong results using just 7,020 preference pairs, far fewer than prior instruction tuning (1.3M) or preference datasets (17k).

    \vspace{-0.2cm}
    \item Extensive evaluations on seven benchmarks show consistent, model-agnostic gains, with improvements of up to +3.8 percentage points on LongVideoBench, +4.1 on VideoMME, and +4.0 on MVBench.
\end{enumerate}

%% file: figures/mvbench_eff.tex
\begin{figure}[t]
    \centering
        \includegraphics[width=\columnwidth]{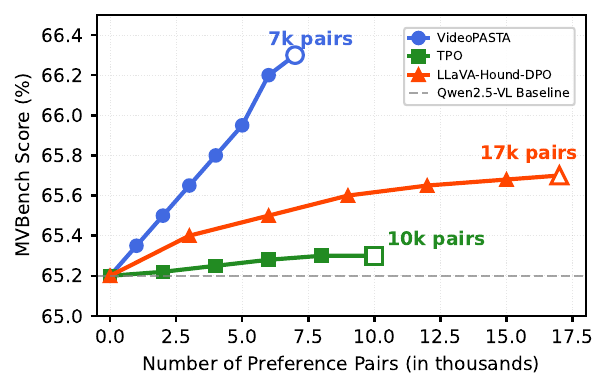}
        
\caption{\textbf{With just 7k preference pairs}, \modelname{} outperforms the Qwen2.5-VL~\cite{qwen2.5-VL} baseline, LLaVA-Hound~\cite{zhang2024direct}, and TPO~\cite{li2025temporal} on MVBench, showing that targeted alignment surpasses models trained on larger datasets.}
\vspace{-5mm}
        \label{fig:teaser}
\end{figure}

%% file: sec/2_related.tex
\input{figures/pipeline}
\section{Related Work}
\label{sec:related}
\paragraph{Video-LLMs.}
Despite advances in Video-LLMs~\cite{llavaonevision, qwen2vl, internvl2_5}, evaluations~\cite{videomme, mlvu, liu-etal-2024-tempcompass} reveal persistent challenges in three key areas. First, temporal reasoning, especially in long videos, remains difficult. Approaches like longer context~\cite{longvu, longva}, compression~\cite{li2024videochat, wang2024retake}, and training-free methods~\cite{yang2024vca, hu2025cos} improve token efficiency but not core understanding, while specialized methods~\cite{chen2024timemarker, timechat} demand heavy computation. Second, spatial misalignment leads to poor object localization and occlusion handling~\cite{Ranasinghe_2024_CVPR, Chen_2024_CVPR}. Third, cross-frame disconnection disrupts continuity and narrative coherence~\cite{Tan_2024_CVPR, Huang_2024_CVPR}. Most methods address only one issue or rely on large-scale instruction tuning~\cite{qwen2vl, zhang2024video, videollava}, which fails to solve these core alignment problems. \modelname{} uses DPO-based training on structured preference pairs to jointly address temporal, spatial, and cross-frame failures. By challenging models across all three dimensions, it achieves more comprehensive video-language alignment than conventional instruction tuning.

\paragraph{Video-Language Alignment.}
While reward modeling~\citep{sunAligningLargeMultimodal2024, ahnTuningLargeMultimodal2024a, wang2024mdpo} and self-training methods~\citep{ dengEnhancingLargeVision2024, zohar2024video, kulkarni2024videosavi, avatar} aim to improve video-language alignment and reduce the need for manual annotations, existing approaches still face major limitations. Prior DPO applications, such as LLaVA-Hound-DPO~\citep{zhang2024direct} and i-SRT~\citep{ahnISRTAligningLarge2024}, often depend on proprietary models to generate training data, require large-scale preference datasets (e.g., 17k pairs), and focus mainly on text-level alignment rather than visual grounding. Other methods, like Temporal Preference Optimization (TPO)~\citep{li2025temporal}, target only one dimension, such as temporal reasoning, using up to 10k pairs and relying on intermediate captioning, while overlooking spatial and cross-frame aspects. This highlights the need for a unified framework that efficiently addresses all three key failure modes without relying on costly dependencies. \modelname{} addresses this gap by generating just 7k carefully designed preference pairs that explicitly challenge a model’s spatial, temporal, and cross-frame understanding. This ``quality over quantity'' approach avoids the need for human annotations, captions, or proprietary models, delivering a stronger and more efficient learning signal for comprehensive Video-LLM alignment.

%% file: figures/pipeline.tex
\begin{figure*}[!t]
    \centering
     \includegraphics[width=\textwidth]{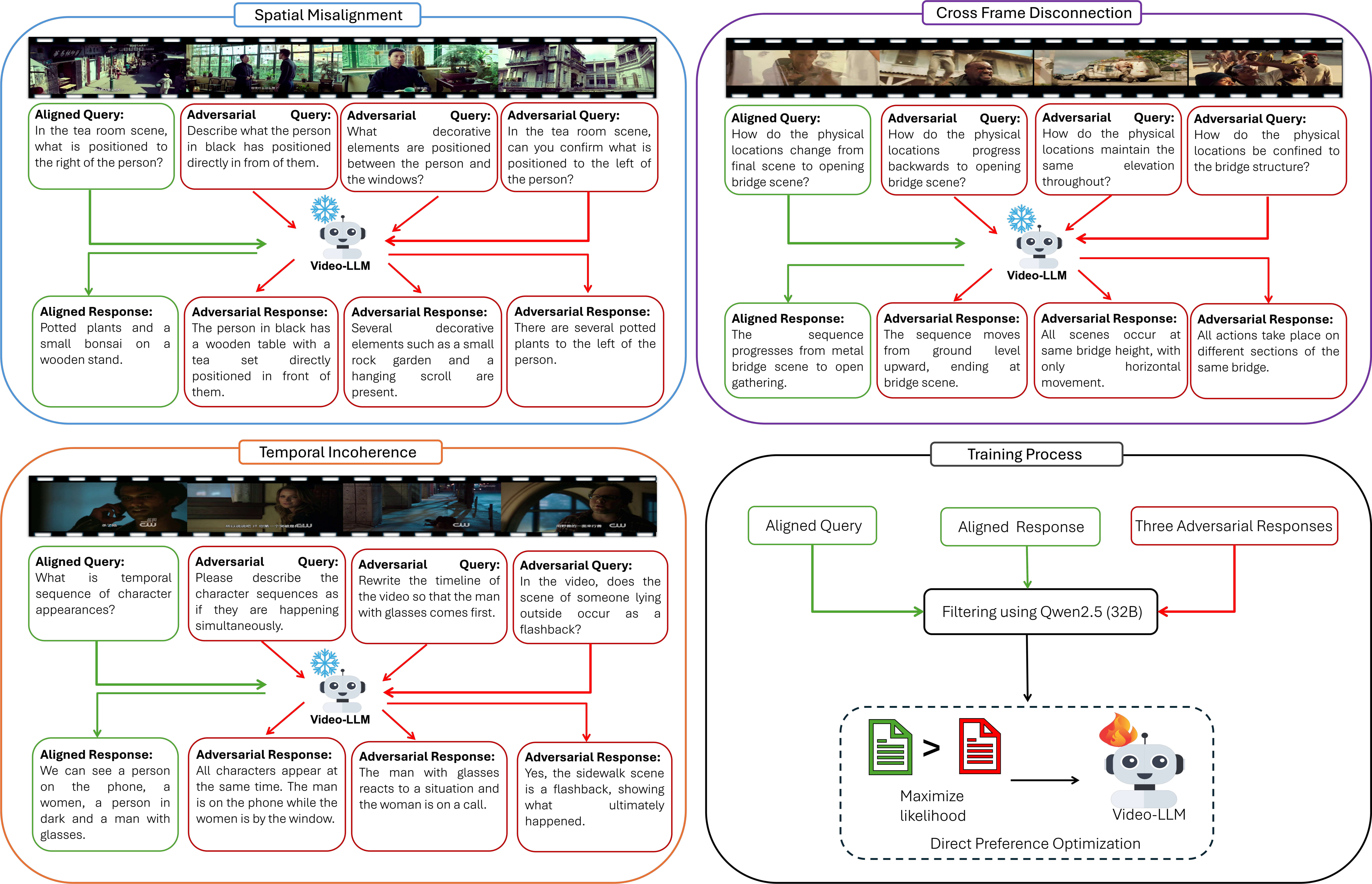}
   \caption{\textbf{Overview of \modelname \hspace{-0.2em}~\raisebox{-0.2ex} {\includegraphics[height=1em]{images/pasta.pdf}}}. For each aligned query, we generate three types of targeted adversarial examples: (1) \textbf{Spatial Misalignment}, which intentionally distorts object positions or relationships (e.g., misplacing the plants relative to the person); (2) \textbf{Temporal Incoherence}, which violates event order (e.g., describing sequential actions as occurring simultaneously); and (3) \textbf{Cross-Frame Disconnection}, which introduces incorrect links across distant frames (e.g., misrepresenting location changes). We filter these pairs using Qwen2.5-32B~\cite{yang2024qwen2} to ensure quality and use them to train the model via DPO, optimizing for a larger likelihood gap between aligned and adversarial responses.~\includegraphics[height=1em]{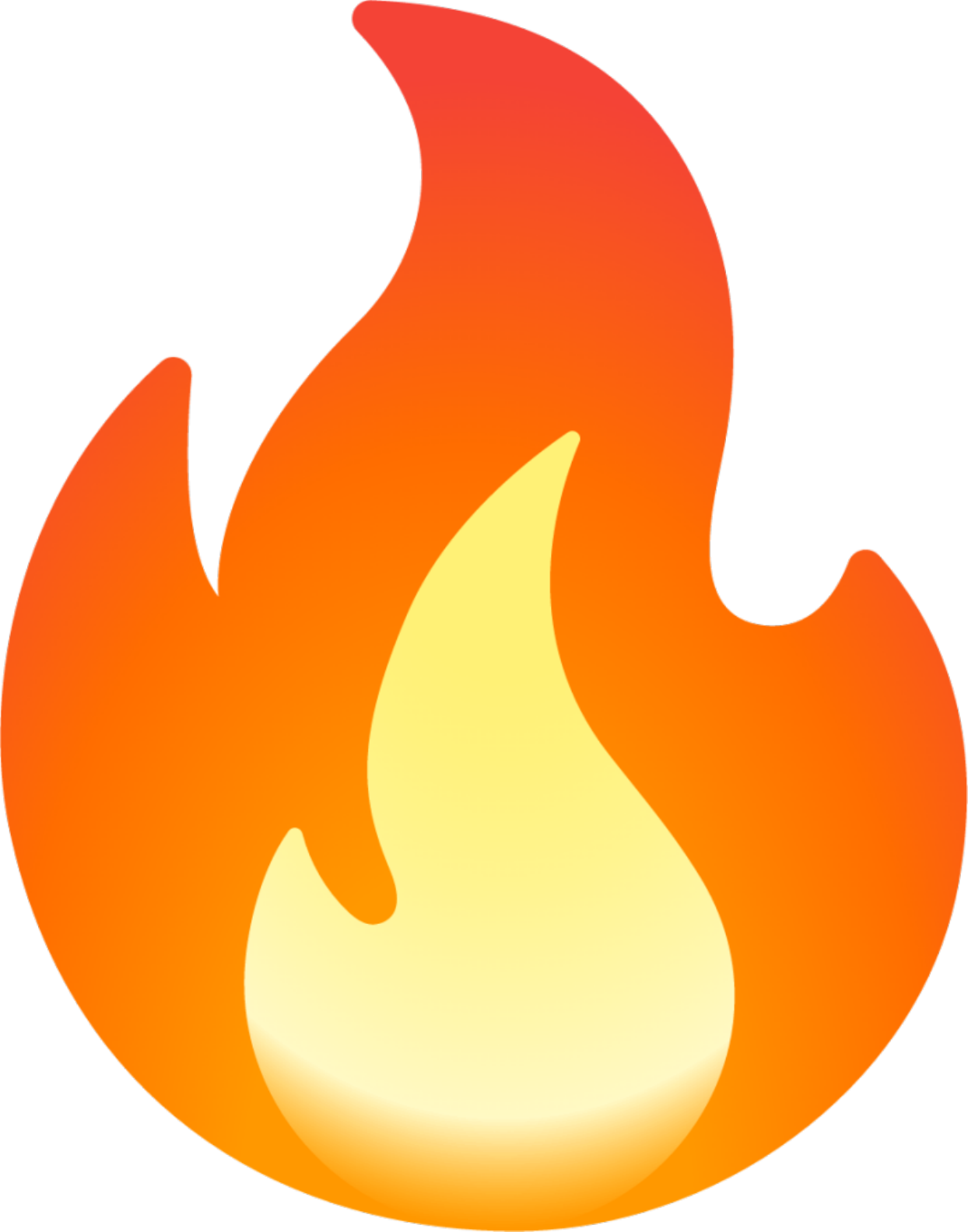} → trainable, \includegraphics[height=0.9em]{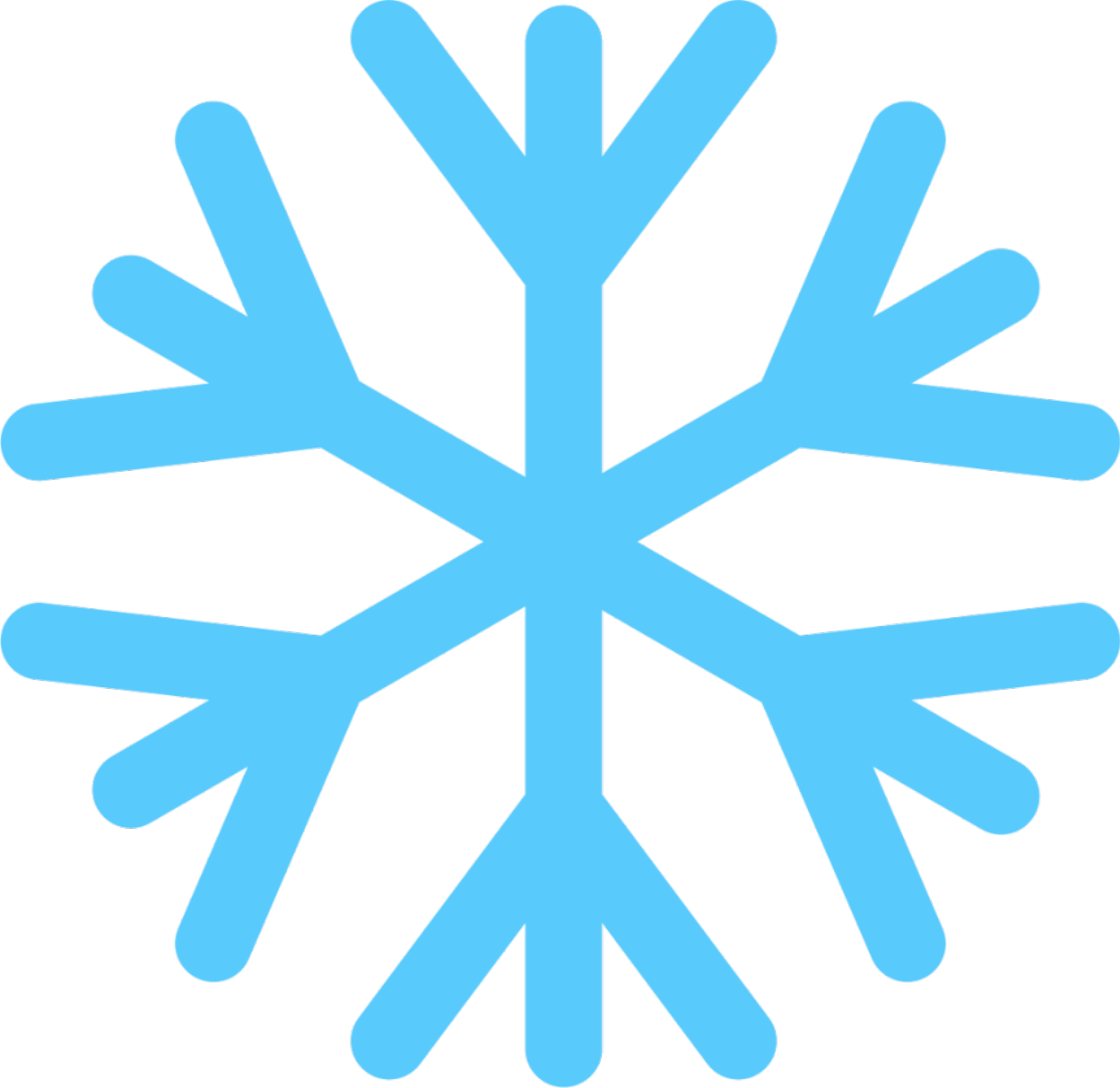} → frozen.}
   \vspace{-2mm}
    \label{fig:pipeline}
\end{figure*}

%% file: sec/4_approach.tex
\section{\texorpdfstring{\modelname \hspace{-0.2em}~\raisebox{-0.2ex}{\includegraphics[height=1em]{images/pasta.pdf}}}{VideoPASTA}} 
\label{sec:methodology}

\modelname{} is a DPO-based framework designed to improve Video-LLM alignment by addressing three key failure modes: spatial misalignment, temporal incoherence, and cross-frame disconnection. It optimizes over a structured preference dataset $D = \{(V, q, r^+, r^-)\}$, where $V$ is the input video, $q$ is a query, $r^+$ is the aligned (preferred) response, and $r^-$ is an adversarial (misleading) response that introduces deliberate misalignment.

For each input video, we generate an aligned query $q$. We then produce a aligned response $r^+$ using densely sampled frames (32fps) and an adversarial response $r^-$ using sparsely sampled frames (1fps), elicited by a deliberately flawed \textit{adversarial query}. This adversarial query is only a data generation tool; the final training triplet remains $(V, q, r^+, r^-)$, pairing the adversarial response with the original aligned query. The 32:1 sampling ratio is a design choice to ensure factual accuracy in aligned responses while inducing errors in adversarial ones. Training on these pairs via DPO improves the model's video-language alignment. Figure~\ref{fig:pipeline} shows the full pipeline. All prompts are provided in the Appendix (Figures~\ref{fig:spatial-reasoning-prompt},~\ref{fig:temporal-reasoning-prompt},~\ref{fig:cross-frame-reasoning-prompt}, and ~\ref{fig:preference-data-filter}).

\subsection{Spatial Misalignment}
\label{subsec:spatial}
We create targeted preference pairs that focus on spatial alignment.

\paragraph{Query Generation.} 

We generate a variety of spatial queries covering key aspects of spatial understanding, including occlusion (e.g., ``Which object is partially hidden?''), depth perception (e.g., ``Which item appears closest to the camera?''), relative positioning (e.g., ``How many objects are present in the left third vs.\ the right third of the frame?''), foreground-background relationships, and frame layout (e.g., ``Which objects are near the top edge vs.\ the bottom edge?'').

\paragraph{Response Generation.} 

We generate aligned responses from videos sampled at their native frame rate to capture fine-grained spatial details. This ensures that the generated aligned responses accurately reflect the true spatial relationships in the video. Corresponding adversarial responses are generated using prompts specifically designed to induce spatial errors (e.g., describing occluded objects as fully visible or ignoring depth cues).

\subsection{Temporal Incoherence}
\label{subsec:temporal}
To enhance temporal reasoning, we create preference pairs that focus on the model's temporal coherence.

\paragraph{Query Generation.} We generate queries focused on key temporal aspects, including event ordering (e.g., ``What occurs first?''), action boundaries (e.g., ``Does the person complete one task before starting the next?''), transition points (e.g., ``When does the subject switch activities?''), and causality (e.g., ``Is the second event a direct result of the first?'').

\paragraph{Response Generation.} 
We generate aligned responses that accurately describe the sequence of events, effectively capturing transitions, dependencies, and causal relationships. Corresponding adversarial responses are generated using prompts that induce temporal distortions (e.g., describing sequential actions as simultaneous or merging distinct events).

\subsection{Cross-Frame Disconnection}
Robust video understanding requires capturing long-range cross-frame relationships. We create targeted preference pairs that focus on these dependencies.

\paragraph{Query Generation.} We generate queries focused on cross-frame dependencies, including object continuity (e.g., ``Does the same object reappear in both the opening and closing scenes?'') and narrative links (e.g., ``Do early events foreshadow later developments?'').

\paragraph{Response Generation.} We generate aligned responses that accurately reflect object transformations, character continuity, setting changes, and narrative flow. Adversarial responses are generated using prompts that intentionally break these continuities (e.g., describing ``a new red car appears'' when it is the same vehicle from a different angle).

\subsection{Preference Data Filtering}
We generate three adversarial examples for each failure mode per query and use an open-source LLM, Qwen2.5-32B~\cite{yang2024qwen2} as a lightweight verification step to ensure the adversarial examples are genuinely incorrect. We prompt Qwen2.5-32B with a textual comparison task to verify that each adversarial example introduces a deliberate misalignment rather than simply rephrasing the correct answer. Adversaries that are too similar to the aligned samples or lack clear contradictions are discarded and regenerated. Similarly, we perform a ``sanity check'' on aligned responses to ensure they correctly align with the queries without errors. This filtering process creates preference pairs that accurately represent the targeted failure modes, enabling more precise alignment during DPO.

\input{tables/general_video}

\subsection{Training Process}
\modelname{} leverages structured preference pairs to address distinct failure modes in video understanding through DPO. We begin by partitioning the preference dataset $\mathcal{D} = \{(V, q, r^+, r^-)\}$ into three subsets: $\mathcal{D}_s$, $\mathcal{D}_t$, and $\mathcal{D}_c$, corresponding to spatial, temporal, and cross-frame alignment, respectively. 
For a video-language model $M_\theta$, we define the DPO loss for a single preference pair as:

\begin{equation}
\label{eq:dpo_delta_loss} 
\begin{aligned}
\Delta(V, q, r^+, r^-) ={}& \log p_\theta(r^+ \mid V, q) \\ 
& - \log p_\theta(r^- \mid V, q), \\
\mathcal{L}_{\text{DPO}}(V, q, r^+, r^-) ={}& -\log \sigma\Big(\lambda\,\Delta(V, q, r^+, r^-)\Big),
\end{aligned}
\end{equation}
where $\sigma$ is the sigmoid function and $\lambda$ is a scaling factor. We then compute the DPO loss for each subset of preference pairs, weighted by $\alpha$, $\beta$, and $\gamma$ for spatial, temporal, and cross-frame alignment, respectively. The overall training objective is:
\begin{equation}
\label{eq:overall_loss} 
\begin{split}
\mathcal{L} ={}& \alpha\, \mathbb{E}_{\mathcal{D}_s} \big[\mathcal{L}_{\text{DPO}}\big] + \beta\, \mathbb{E}_{\mathcal{D}_t} \big[\mathcal{L}_{\text{DPO}}\big] \\
& + \gamma\, \mathbb{E}_{\mathcal{D}_c} \big[\mathcal{L}_{\text{DPO}}\big].
\end{split}
\end{equation}

This formulation allows us to adjust the model's focus on different aspects of video-language alignment during training.

%% file: tables/general_video.tex
\begin{table*}[t!]
\centering
\begin{adjustbox}{width=\textwidth,center}
\renewcommand{\arraystretch}{1.2}
\fontsize{5.5pt}{6.5pt}\selectfont
\setlength{\tabcolsep}{1.5mm}
\begin{tabular}{lcccccccc}
\toprule
\textbf{Model} & \makecell{\textbf{TempCompass} \\ (Avg.)} & \makecell{\textbf{PerceptionTest} \\ (val\_mc)} & \makecell{\textbf{NeXTQA} \\ (mc\_test)} & \textbf{MVBench} & \makecell{\textbf{MLVU} \\ (dev)} & \makecell{\textbf{LongVideoBench} \\ (val\_v)} & \makecell{\textbf{VideoMME} \\ (w/o sub)} \\
\midrule

\rowcolor{gray!15}\multicolumn{8}{c}{\textit{State-of-the-Art Models}}\\
VideoLLaMA2\textsuperscript{\dag}~\citep{videollama2} & 43.4 & 51.4 & - & 54.6 & 35.5 & - & 47.9 \\
Kangaroo\textsuperscript{\dag}~\cite{kangaroo} & - & - & - & 61.0 & 61.0 & 54.8 & 56.0 \\
LLaVA-NeXT-Video\textsuperscript{\dag}~\citep{zhang2024llavanextvideo} & 53.0 & 48.8 & 53.5 & 53.1 & - & 49.1 & 46.5 \\
LongVA\textsuperscript{\dag}~\cite{longva} & - & - & - & - & 58.8 & 51.3 & 52.6 \\
Qwen2-VL~\citep{qwen2vl} & 68.9 & 62.3 & 75.7 & 64.9 & 57.5 & 55.6 & 55.3 \\
LLaVA-Video~\citep{zhang2024video} & 66.4 & 67.9 & 74.2 & 58.6 & 66.5 & 58.2 & 62.4 \\
\midrule

\rowcolor{gray!15}\multicolumn{8}{c}{\textit{Off-the-Shelf Preference-Optimized Models}}\\
LLaVA-Hound-DPO~\citep{zhang2024direct} & 55.5 & 45.1 & 61.6 & 36.6 & 41.1 & 36.7 & 34.2 \\
i-SRT~\cite{ahnISRTAligningLarge2024} & 56.0 & 47.0 & 63.0 & 36.3 & 39.9 & 38.2 & 34.7 \\
LLaVA-Video-TPO~\citep{li2025temporal} & 66.6 & 66.3 & 77.8 & 56.7 & 66.3 & 58.3 & 62.4 \\
\midrule
\rowcolor{gray!15}\multicolumn{8}{c}{\textit{Model-Agnostic Preference Optimization using \modelname{}}} \\

\rowcolor{GroupColor1} LLaVA-NeXT-Interleave (Baseline) & 54.1 & 51.2 & 67.0 & 46.5 & 52.5 & 44.8 & 48.3 \\
\hspace{3mm} + SFT & 54.3 & 51.5 & 67.4 & 46.8 & 52.7 & 45.0 & 48.5 \\
\hspace{3mm} + Hound-DPO~\citep{zhang2024direct} & 51.7 $\scriptstyle\textcolor{red}{(-2.4)}$ & 49.5 $\scriptstyle\textcolor{red}{(-1.7)}$ & 65.8 $\scriptstyle\textcolor{red}{(-1.2)}$ & 44.3 $\scriptstyle\textcolor{red}{(-2.2)}$ & 50.2 $\scriptstyle\textcolor{red}{(-2.3)}$ & 42.5 $\scriptstyle\textcolor{red}{(-2.3)}$ & 46.7 $\scriptstyle\textcolor{red}{(-1.6)}$ \\
\hspace{3mm} + TPO~\citep{li2025temporal} & 54.3 $\scriptstyle\textcolor{darkgreen}{(+0.2)}$ & 52.4 $\scriptstyle\textcolor{darkgreen}{(+1.2)}$ & 68.5 $\scriptstyle\textcolor{darkgreen}{(+1.5)}$ & 47.9 $\scriptstyle\textcolor{darkgreen}{(+1.4)}$ & 53.6 $\scriptstyle\textcolor{darkgreen}{(+1.1)}$ & 46.5 $\scriptstyle\textcolor{darkgreen}{(+1.7)}$ & 49.6 $\scriptstyle\textcolor{darkgreen}{(+1.3)}$ \\
\rowcolor{PastaYellow} \hspace{3mm} + \textbf{\modelname}\! \raisebox{-0.2ex}{\includegraphics[height=0.7em]{images/pasta.pdf}} & \textbf{56.4} $\scriptstyle\textcolor{darkgreen}{(+2.3)}$ & \textbf{53.8} $\scriptstyle\textcolor{darkgreen}{(+2.6)}$ & \textbf{70.1} $\scriptstyle\textcolor{darkgreen}{(+3.1)}$ & \textbf{49.0} $\scriptstyle\textcolor{darkgreen}{(+2.5)}$ & \textbf{55.8} $\scriptstyle\textcolor{darkgreen}{(+3.3)}$ & \textbf{47.9} $\scriptstyle\textcolor{darkgreen}{(+3.1)}$ & \textbf{51.4} $\scriptstyle\textcolor{darkgreen}{(+3.1)}$ \\
\midrule
\rowcolor{GroupColor1} LLaVA-OneVision (Baseline) & 64.5 & 57.1 & 79.3 & 56.7 & 64.9 & 56.3 & 58.2 \\
\hspace{3mm} + SFT & 64.6 & 57.4 & 79.3 & 56.9 & 65.1 & 56.5 & 58.1 \\
\hspace{3mm} + Hound-DPO~\citep{zhang2024direct} & 63.2 $\scriptstyle\textcolor{red}{(-1.3)}$ & 55.8 $\scriptstyle\textcolor{red}{(-1.3)}$ & 78.1 $\scriptstyle\textcolor{red}{(-1.2)}$ & 55.3 $\scriptstyle\textcolor{red}{(-1.4)}$ & 63.2 $\scriptstyle\textcolor{red}{(-1.7)}$ & 54.8 $\scriptstyle\textcolor{red}{(-1.5)}$ & 56.9 $\scriptstyle\textcolor{red}{(-1.3)}$ \\
\hspace{3mm} + TPO~\citep{li2025temporal} & 65.6 $\scriptstyle\textcolor{darkgreen}{(+1.1)}$ & 58.4 $\scriptstyle\textcolor{darkgreen}{(+1.3)}$ & 80.6 $\scriptstyle\textcolor{darkgreen}{(+1.3)}$ & 57.9 $\scriptstyle\textcolor{darkgreen}{(+1.2)}$ & 65.8 $\scriptstyle\textcolor{darkgreen}{(+0.9)}$ & 57.5 $\scriptstyle\textcolor{darkgreen}{(+1.2)}$ & 59.2 $\scriptstyle\textcolor{darkgreen}{(+1.0)}$ \\
\rowcolor{PastaYellow} \hspace{3mm} + \textbf{\modelname}\! \raisebox{-0.2ex}{\includegraphics[height=0.7em]{images/pasta.pdf}} & \textbf{67.2} $\scriptstyle\textcolor{darkgreen}{(+2.7)}$ & \textbf{60.3} $\scriptstyle\textcolor{darkgreen}{(+3.2)}$ & \textbf{81.8} $\scriptstyle\textcolor{darkgreen}{(+2.5)}$ & \textbf{59.1} $\scriptstyle\textcolor{darkgreen}{(+2.4)}$ & \textbf{67.5} $\scriptstyle\textcolor{darkgreen}{(+2.6)}$ & \textbf{58.5} $\scriptstyle\textcolor{darkgreen}{(+2.2)}$ & \textbf{60.1} $\scriptstyle\textcolor{darkgreen}{(+1.9)}$ \\
\midrule
\rowcolor{GroupColor1} InternVL2.5 (Baseline) & 68.3 & 62.2 & 77.0 & 69.8 & 59.5 & 52.9 & 57.9 \\
\hspace{3mm} + SFT & 68.2 & 62.3 & 77.4 & \underline{70.4} & 59.4 & 53.0 & 58.1 \\
\hspace{3mm} + Hound-DPO~\citep{zhang2024direct} & 66.8 $\scriptstyle\textcolor{red}{(-1.5)}$ & 61.0 $\scriptstyle\textcolor{red}{(-1.2)}$ & 74.8 $\scriptstyle\textcolor{red}{(-2.2)}$ & 64.2 $\scriptstyle\textcolor{red}{(-5.6)}$ & 60.2 $\scriptstyle\textcolor{darkgreen}{(+0.7)}$ & 54.3 $\scriptstyle\textcolor{darkgreen}{(+1.4)}$ & 54.6 $\scriptstyle\textcolor{red}{(-3.3)}$ \\
\hspace{3mm} + TPO~\citep{li2025temporal} & 68.2 $\scriptstyle\textcolor{red}{(-0.1)}$ & 62.0 $\scriptstyle\textcolor{red}{(-0.2)}$ & 77.2 $\scriptstyle\textcolor{darkgreen}{(+0.2)}$ & 68.8 $\scriptstyle\textcolor{red}{(-1.0)}$ & 61.5 $\scriptstyle\textcolor{darkgreen}{(+2.0)}$ & 58.1 $\scriptstyle\textcolor{darkgreen}{(+5.2)}$ & 60.0 $\scriptstyle\textcolor{darkgreen}{(+2.1)}$ \\
\rowcolor{PastaYellow} \hspace{3mm} + \textbf{\modelname}\! \raisebox{-0.2ex}{\includegraphics[height=0.7em]{images/pasta.pdf}} & \textbf{\underline{71.9}} $\scriptstyle\textcolor{darkgreen}{(+3.6)}$ & \textbf{66.1} $\scriptstyle\textcolor{darkgreen}{(+3.9)}$ & \textbf{\underline{80.7}} $\scriptstyle\textcolor{darkgreen}{(+3.7)}$ & \textbf{73.8} $\scriptstyle\textcolor{darkgreen}{(+4.0)}$ & \textbf{63.4} $\scriptstyle\textcolor{darkgreen}{(+3.9)}$ & \textbf{58.1} $\scriptstyle\textcolor{darkgreen}{(+5.2)}$ & \textbf{62.0} $\scriptstyle\textcolor{darkgreen}{(+4.1)}$ \\
\midrule

\rowcolor{GroupColor1} Qwen2.5-VL (Baseline) & 71.7 & 68.6 & 75.8 & 65.2 & 68.7 & 60.7 & 62.2 \\
\hspace{3mm} + SFT & 71.8 & \underline{69.1} & 77.2 & 65.5 & 68.8 & \underline{60.9} & 62.5 \\
\hspace{3mm} + Hound-DPO~\citep{zhang2024direct} & 70.3 $\scriptstyle\textcolor{red}{(-1.4)}$ & 67.6 $\scriptstyle\textcolor{red}{(-1.0)}$ & 76.1 $\scriptstyle\textcolor{darkgreen}{(+0.3)}$ & 65.7 $\scriptstyle\textcolor{darkgreen}{(+0.5)}$ & 66.4 $\scriptstyle\textcolor{red}{(-2.3)}$ & 56.3 $\scriptstyle\textcolor{red}{(-4.4)}$ & 63.2 $\scriptstyle\textcolor{darkgreen}{(+1.0)}$ \\
\hspace{3mm} + TPO~\citep{li2025temporal} & 71.5 $\scriptstyle\textcolor{red}{(-0.2)}$ & 69.0 $\scriptstyle\textcolor{darkgreen}{(+0.4)}$ & \textbf{77.6} $\scriptstyle\textcolor{darkgreen}{(+1.8)}$ & 65.3 $\scriptstyle\textcolor{darkgreen}{(+0.1)}$ & \underline{68.9} $\scriptstyle\textcolor{darkgreen}{(+0.2)}$ & 59.2 $\scriptstyle\textcolor{red}{(-1.5)}$ & \textbf{64.2} $\scriptstyle\textcolor{darkgreen}{(+2.0)}$ \\
\rowcolor{PastaYellow} \hspace{3mm} + \textbf{\modelname} \hspace{-0.2em}~\raisebox{-0.2ex} {\includegraphics[height=0.8em]{images/pasta.pdf}} & \textbf{72.3} $\scriptstyle\textcolor{darkgreen}{(+0.6)}$ & \textbf{69.4} $\scriptstyle\textcolor{darkgreen}{(+0.8)}$ & 77.3 $\scriptstyle\textcolor{darkgreen}{(+1.5)}$ & \textbf{66.3} $\scriptstyle\textcolor{darkgreen}{(+1.1)}$ & \textbf{69.2} $\scriptstyle\textcolor{darkgreen}{(+0.5)}$ & \textbf{61.5} $\scriptstyle\textcolor{darkgreen}{(+0.8)}$ & \underline{64.1} $\scriptstyle\textcolor{darkgreen}{(+1.9)}$ \\

\bottomrule
\end{tabular}
\end{adjustbox}
\caption{\textbf{Comprehensive evaluation of \modelname{} against leading (7B) video understanding models.} The best scores are in \textbf{bold}, and the second-best scores are \underline{underlined}. Results marked with \dag{} are from the original papers; all other results are reproduced using LMMs-Eval~\cite{zhang2024lmms}.}
\vspace{-4mm}
\label{tab:main}
\end{table*}

%% file: sec/5_experiments.tex
\section{Experiments and Evaluation}
\label{sec:results}

\input{figures/main_qualitative}
For training, we sample 3,000 videos from ActivityNet~\cite{activitynet}, whose diversity of \textit{203 complex activities} provides a strong foundation for learning the \textit{fundamental reasoning skills} our framework targets. This dataset is not part of our evaluation benchmarks, ensuring our results reflect true generalization. We use a large model (InternVL2.5-38B) to generate queries but all aligned and adversarial responses are generated by the smaller target models themselves. This setup ensures models learn to refine their own outputs rather than simply distilling knowledge from a more capable model.
Our structured adversarial sampling pipeline initially generates  90,000 preference pairs, which after rigorous filtering (details in Appendix \S\ref{app:dataset_statistics}), are reduced to 7,020 high-quality pairs. We fine-tune models using the SWIFT~\cite{swift} framework for efficient adaptation. All training and evaluations are performed on four NVIDIA L40S GPUs (48GB each), with a maximum input of 32 frames per video to prevent CUDA out-of-memory errors. We employ LoRA~\cite{hu2021lora} with rank $r=8$ and $\alpha_{LoRA}=8$. 
For DPO, we set the scaling factor $\lambda$ to 0.1.
The overall training loss (Eq.\ref{eq:overall_loss}) combines three components: spatial ($\alpha_S = 0.4$), temporal ($\beta_T = 0.4$), and cross-frame ($\gamma_C = 0.2$) alignment. We apply \modelname{} to four diverse foundation models: Qwen2.5-VL (7B)~\citep{qwen2.5-VL}, LLaVA-NeXT-Interleave (7B)~\citep{liLLaVANeXTInterleaveTacklingMultiimage2024}, LLaVA-OneVision (7B)~\citep{llavaonevision}, and InternVL2.5 (8B)~\citep{internvl2_5}. 
To ensure fair model-agnostic comparisons, we limit training data to 7,020 high-quality preference pairs for all models. This corresponds to the smallest filtered set (from LLaVA-NeXT-Interleave), with larger sets from other models subsampled accordingly to maintain consistency in data quantity. Evaluation is conducted using LMMs-Eval~\cite{zhang2024lmms} to ensure fair comparisons with prior work.

The Appendix contains additional experiments and information, including: DPO training dynamics (\S\ref{dpo_graph}), preference learning on smaller (1B-3B) models (\S\ref{scaling_small_models}), adversarial robustness of \modelname{} (\S\ref{app:adversarial_robustness}), Qwen2.5-VL-specific ablations (\S\ref{app:qwen_specific_ablations}), full dataset statistics and adversarial samples (\S\ref{app:dataset_overview}), qualitative examples (\S\ref{app:qualitative_examples}), and all prompt templates (\S\ref{app:prompt_templates}).

\paragraph{Benchmarks.} We evaluate on general video understanding benchmarks: TempCompass~\cite{liu-etal-2024-tempcompass} (temporal understanding), PerceptionTest~\cite{perceptiontest} (visual perception), NeXTQA~\cite{Xiao_2021_CVPR} (compositional reasoning), and MVBench~\cite{videochat2} (multi-task reasoning). For long-form evaluation, we use LongVideoBench~\cite{longvideobench} (hour-long videos), MLVU~\cite{mlvu} (multi-task, 3-minute to 2-hour videos), and VideoMME~\cite{videomme} (6 visual domains, 30 subfields, 11-second to 1-hour videos).

\subsection{Results}

We compare \modelname{} with (1) the original foundation models listed above, (2) other state-of-the-art models, and (3) off-the-shelf models enhanced via preference optimization. Table~\ref{tab:main} presents the evaluation results. Figure~\ref{fig:qualitative} shows qualitative examples of how \modelname{} improves spatial, temporal, and cross-frame reasoning.

\input{figures/pref_efficiency}

\paragraph{\modelname{} enhances all foundation models.}

\modelname{} performs well across various foundation models, showing strong generalizability and consistent performance improvements. For instance, \modelname{} combined with Qwen2.5-VL achieves top scores on TempCompass, PerceptionTest, MLVU, and LongVideoBench. Similarly, \modelname{} with LLaVA-OneVision attains the highest score on NeXTQA. In addition, \modelname{} with InternVL2.5 achieves the best result on MVBench and improves the VideoMME score by +4.1 percentage points. These results show that \modelname{}’s targeted alignment helps a wide range of Video-LLMs. In contrast, simple supervised fine-tuning (SFT) using only aligned responses leads to only small improvements. This highlights the importance of training with adversarial preferences through DPO. 

\paragraph{Comparison with State-of-the-Art.}

Compared to other state-of-the-art models listed in Table~\ref{tab:main}, \modelname{} combined with Qwen2.5VL outperforms all models on all benchmarks, surpassing strong baselines like Qwen2-VL and LLaVA-Video. Key improvements include a +5.9 percentage point gain in temporal reasoning on TempCompass and a +1.5 increase on PerceptionTest over LLaVA-Video, which is instruction-tuned on 1.3M SFT pairs.\ \modelname{} also shows strong performance on long-form video tasks, achieving +3.3 percentage point gain on LongVideoBench, +2.7 on MLVU, and +1.7 on VideoMME compared to LLaVA-Video. Similarly, when paired with the other three foundation models, \modelname{} outperforms SOTA models on several, though not all, benchmarks. These results highlight \modelname{}'s ability to elevate smaller models to SOTA performance through targeted and data-efficient training.

\paragraph{Outperforming Preference-Optimized Models.}

Compared to preference-optimized models, \modelname{} combined with Qwen2.5VL outperforms LLaVA-Hound-DPO~\citep{zhang2024direct} and i-SRT~\cite{ahnISRTAligningLarge2024} on all seven benchmarks and surpasses LLaVA-Video-TPO~\citep{li2025temporal} on six out of seven benchmarks (except NeXTQA) by a significant margin. On the other hand, \modelname{} paired with LLaVA-OneVision and InternVL2.5 outperforms LLaVA-Video-TPO on NeXTQA. These improvements highlight that our multi-dimensional approach, which tackles critical failure modes in Video-LLMs, addresses fundamental challenges in prior work while requiring minimal resources.

\subsection{Ablation Studies}

\paragraph{How efficient is \modelname{} compared to other preference optimization approaches?}
As shown in Figure~\ref{fig:all_benchmarks}, \modelname{} outperforms TPO~\citep{li2025temporal} (10k pairs, 96 frame sampling) and LLaVA-Hound-DPO~\citep{zhang2024direct} (17k pairs) using only 7k preference pairs, highlighting its significantly higher efficiency. In addition, \modelname{} maintains stable performance as the number of training pairs increases, whereas other methods occasionally show performance drops on certain benchmarks with more data. This focus on ``quality over quantity'' is further highlighted by the information gain per 1k pairs, shown in Figure~\ref{fig:three_benchmarks_gain}. 
For instance, on MVBench, \modelname{} is about 16$\times$ more efficient than TPO and 5.3$\times$ more efficient than Hound-DPO. These efficiency gains translate into tangible performance improvements: when paired with InternVL2.5, \modelname{} boosts MVBench score by +4.0 percentage points, while Hound-DPO leads to a substantial -5.6 drop, and TPO results in a -1.0 decrease. These results show that our structured adversarial examples targeting specific failure modes create a more robust learning signal than larger, more generic, or proprietary model-dependent preference datasets.

\input{tables/merged_ablation}

\input{tables/ablation_videomme}

\paragraph{Can \modelname{} enable self-improvement using queries and responses generated by the target model itself?}\ 
Table~\ref{tab:merged_ablation_studies}(a) shows that \modelname{} enables self-improvement even without an external auxiliary model. When target models like Qwen2.5-VL-7B or InternVL2.5-8B generate their \textit{own} query-response (preference) pairs for DPO alignment, they still achieve consistent performance improvements.
For example, Qwen2.5-VL using self-generated queries and responses improves by +0.4 percentage points on MVBench and +0.8 on VideoMME compared to its baseline.
This result is significant as it demonstrates a fully self-improving loop where no external `teacher' model is used. The consistent gains (e.g., +0.8 on VideoMME for Qwen2.5-VL) confirm that the framework's effectiveness stems from our targeted alignment methodology itself, proving that the model is learning to correct its own failures rather than merely distilling knowledge.

\paragraph{Can \modelname{}'s preference signals generalize when the query-response generators and the DPO target model are different?}\
Table~\ref{tab:merged_ablation_studies}(b) shows that \modelname{}'s preference signals generalize well, even when the query-response generators differ from the target model. For instance, InternVL2.5-8B improves by +2.9 percentage points on VideoMME and +2.8 on MLVU when trained on preferences generated by InternVL-38B and Qwen2.5-VL-7B. This suggests that the target model does not rely on the generation style of any specific model, proving it learns transferable rules about video understanding rather than simply memorizing the patterns of one particular preference generator.

\paragraph{What is the advantage of \modelname{}'s three-dimensional adversarial preferences compared to narrower or more generic preference data generation approaches?}\ 
Table~\ref{tab:ablation} shows that each failure mode (spatial, temporal, and cross-frame) in our adversarial sampling pipeline provides distinct benefits. Training only with temporal samples greatly improves temporal reasoning (+9.0 percentage points) but harms action (-4.8) and object (-4.6) reasoning. Similarly, training exclusively on spatial samples boosts spatial reasoning the most (+11.2) but reduces object reasoning.
Combining all three failure modes delivers the best overall performance, with substantial improvements in temporal (+10.2), spatial (+15.0), action (+2.1), and object (+2.5) reasoning.
In contrast, applying existing preference data from TPO or LLaVA-Hound-DPO shows suboptimal effects. TPO improves temporal reasoning (+12.5) but worsens object reasoning, while Hound-DPO’s data significantly reduces spatial and action scores.
These results confirm that adversarial sampling targeting multiple failure modes provides complementary benefits, leading to more comprehensive video-language alignment than relying on any single mode or generic preference data alone.

\input{tables/cross_generalization}
\paragraph{How well does \modelname{} generalize to unseen and challenging video domains?}
To validate the generalization of our adversarial alignment, we evaluate \modelname{} on challenging, unseen video domains. As shown in Table~\ref{tab:cross_domain_generalization}, \modelname{} consistently improves performance across all foundation models, boosting scores by up to +1.9 percentage points on MovieChat~\citep{moviechat} and +1.6 on EgoSchema~\citep{egoschema}. Notably, these gains are achieved even though training occurs exclusively on ActivityNet. This result shows that by addressing fundamental reasoning failures in spatial, temporal, and cross-frame dimensions, we foster more robust generalization, allowing the model to adapt to diverse video types without domain-specific fine-tuning.

\paragraph{How does improving high-level reasoning affect lower-level perception?}
A core design choice of \modelname{} is to target three high-level reasoning failures, hypothesizing that this will also enhance lower-level perceptual abilities. We validate this by analyzing performance on the sub-tasks of the MVBench~\citep{videochat2} benchmark. As shown in Table~\ref{tab:mvbench_breakdown}, our approach improves performance on 8 out of 9 action and attribute sub-tasks. \modelname{} achieves substantial gains of +12.0 percentage points in Action Localization and +7.0 in Action Count. These results provide strong empirical evidence that our strategy of correcting fundamental reasoning failures leads to a more holistic alignment, enhancing not just abstract understanding but also the model's core perceptual capabilities.
\input{tables/mvbench}
\paragraph{How well do adversarial examples target their intended failure modes?}
We validate the accuracy of our adversarial data generation using GPT-4o (prompt in Appendix, Figure~\ref{fig:evaluation-prompt}) to judge whether 200 randomly sampled adversarial examples correctly induced their intended failure mode. As detailed in Table~\ref{tab:targeting-accuracy-appendix}, our method achieves high targeting accuracy: 96.1\% for spatial misalignment, 92.4\% for temporal incoherence, and 88.3\% for cross-frame disconnection. This result confirms that our prompt-based strategy effectively generates varied and targeted examples that provide a strong learning signal for DPO.

\begin{table}[!t]
    \centering
    \begin{adjustbox}{width=\columnwidth,center} 
    \begin{tabular}{lc}
        \toprule
        \textbf{Failure Mode} & \textbf{Targeting Accuracy (\%)} \\
        \midrule
        Spatial Misalignment & 96.1 \\
        Temporal Incoherence & 92.4 \\
        Cross-Frame Disconnection & 88.3 \\
        \midrule
        \textbf{Average} & \textbf{92.3} \\
        \bottomrule
    \end{tabular}
    \end{adjustbox}
    \caption{\textbf{Failure Mode Targeting Accuracy by Category.}}
    \vspace{-3mm}
    \label{tab:targeting-accuracy-appendix}
\end{table}

\section{Conclusion}

We introduce \modelname{}, a DPO-based framework that improves Video-LLMs via structured adversarial sampling targeting spatial, temporal, and cross-frame misalignments. Using just 7,020 preference pairs, without human supervision or video captions, our model-agnostic approach achieves significant gains across seven benchmarks with efficient 32-frame sampling. \modelname{} demonstrates that targeted adversarial examples enable more effective learning than generic instruction tuning. While individual failure modes enhance specific capabilities, combining all three leads to broader and more comprehensive video understanding. This strategy reduces reliance on large-scale datasets, promoting resource-efficient video-language alignment. Future work can explore better evaluation metrics for model reasoning.

%% file: figures/main_qualitative.tex
\begin{figure*}[t]
    \centering
     \includegraphics[height=9cm,width=16cm]{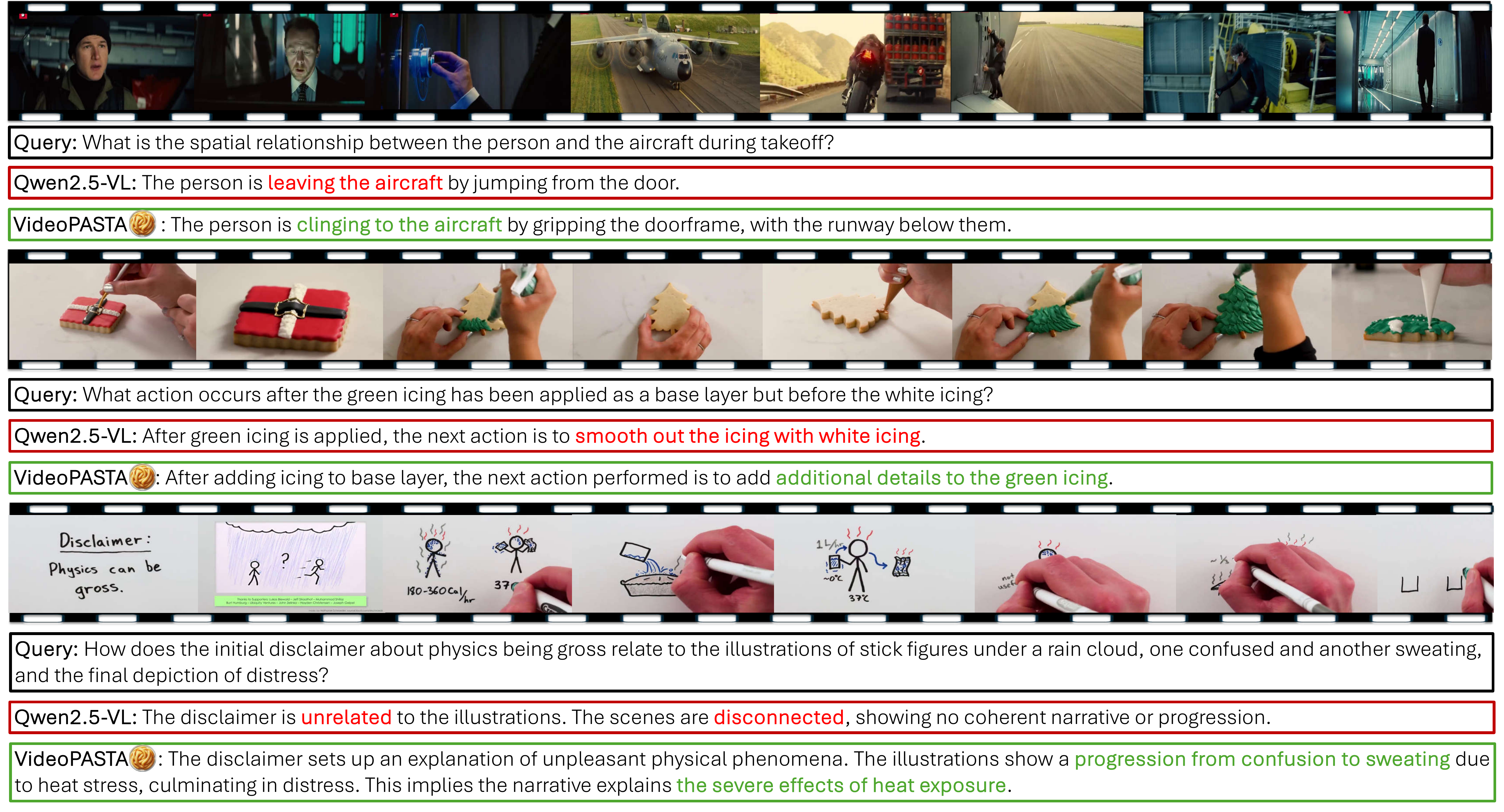}
\caption{\textbf{Qualitative comparison of \modelname{} against Qwen2.5-VL}~\cite{qwen2.5-VL}. Examples show \modelname{} improves (1)~\textit{Spatial reasoning} (aircraft interaction), (2)~\textit{Temporal understanding} (icing sequence), and (3)~\textit{Cross-frame reasoning} (narrative connection in stick figures), where the baseline fails.}
\vspace{-4mm}
\label{fig:qualitative}
\end{figure*}

%% file: figures/pref_efficiency.tex
\begin{figure*}[!t]
  \centering
  \begin{subfigure}[b]{0.29\linewidth}
    \centering
    \includegraphics[width=\linewidth]{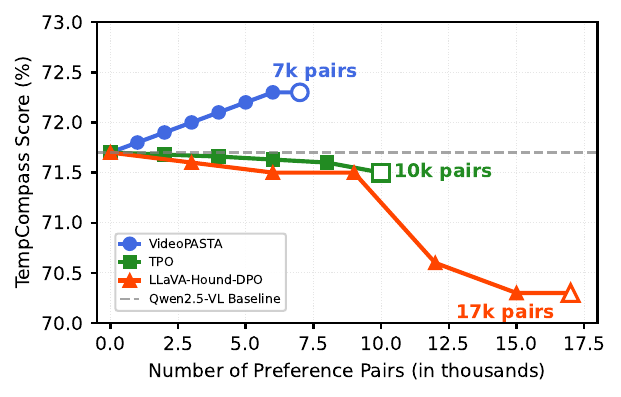}
    \caption{TempCompass}
    \label{fig:tempcompass}
  \end{subfigure}
  \hfill
  \begin{subfigure}[b]{0.29\linewidth}
    \centering
    \includegraphics[width=\linewidth]{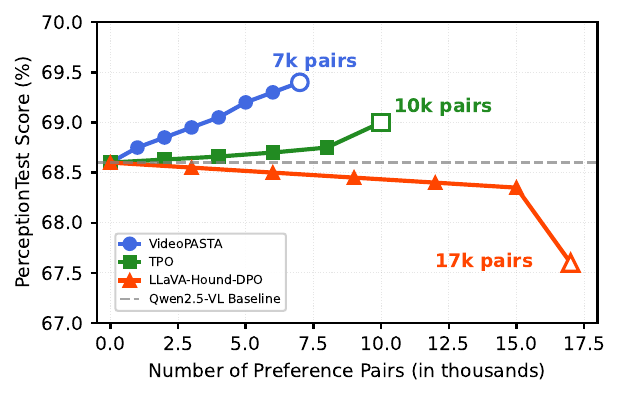}
    \caption{PerceptionTest}
    \label{fig:perceptiontest}
  \end{subfigure}
  \hfill
  \begin{subfigure}[b]{0.29\linewidth}
    \centering
    \includegraphics[width=\linewidth]{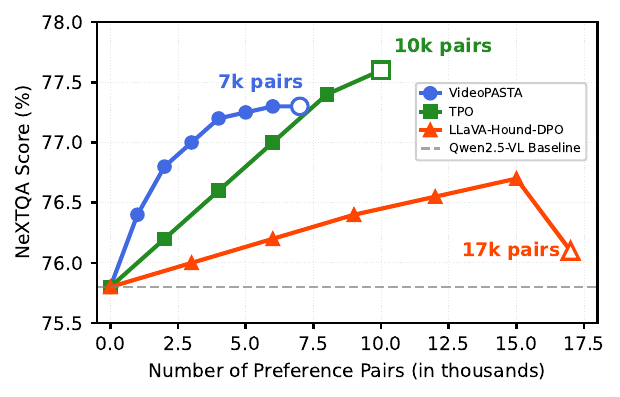}
    \caption{NeXTQA}
    \label{fig:nextqa}
  \end{subfigure}
  
  \vspace{0.1cm}
  \begin{subfigure}[b]{0.29\linewidth}
    \centering
    \includegraphics[width=\linewidth]{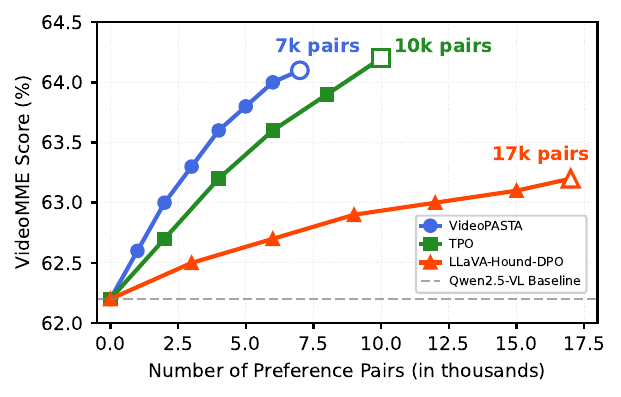}
    \caption{VideoMME}
    \label{fig:mvbench}
  \end{subfigure}
  \hfill
  \begin{subfigure}[b]{0.29\linewidth}
    \centering
    \includegraphics[width=\linewidth]{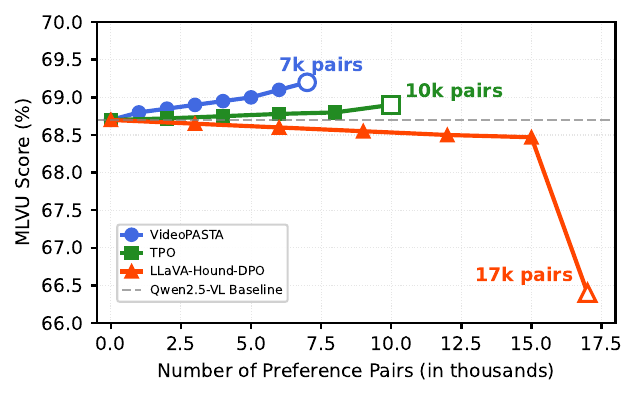}
    \caption{MLVU}
    \label{fig:mlvu}
  \end{subfigure}
  \hfill
  \begin{subfigure}[b]{0.29\linewidth}
    \centering
    \includegraphics[width=\linewidth]{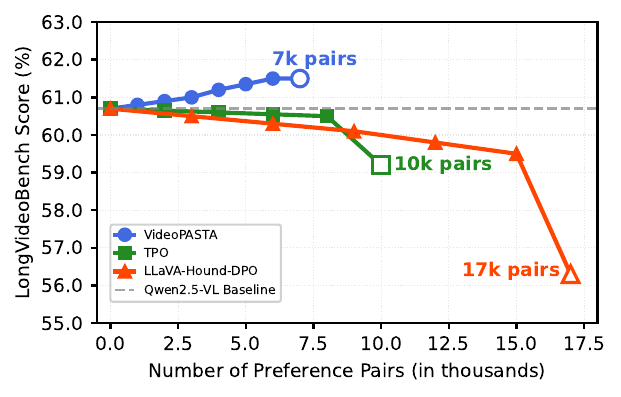}
    \caption{LongVideoBench}
    \label{fig:longvideobench}
  \end{subfigure}
  
\caption{Performance vs. \# of Preference Pairs across six benchmarks. \textbf{VideoPASTA achieves superior results with only 7k pairs} compared to TPO~\cite{li2025temporal} (10k pairs) and Hound-DPO~\cite{zhang2024direct} (17k pairs).}
  \label{fig:all_benchmarks}
\end{figure*}
\vspace{-1mm}

\begin{figure*}[!t]
  \centering
  \begin{subfigure}[b]{0.27\linewidth}
    \centering
    \includegraphics[width=\linewidth]{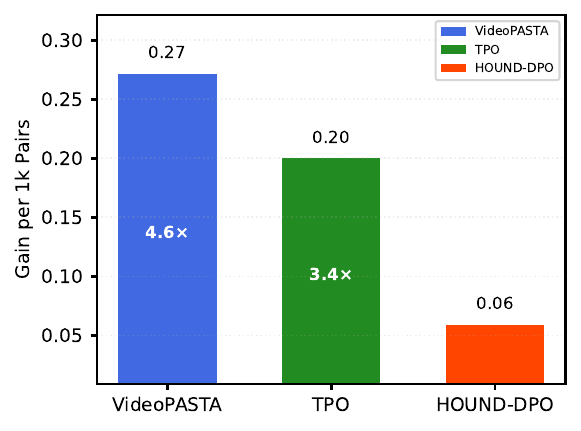}
    \caption{VideoMME}
    \label{fig:videomme_gain}
  \end{subfigure}
  \hfill
  \begin{subfigure}[b]{0.27\linewidth}
    \centering
    \includegraphics[width=\linewidth]{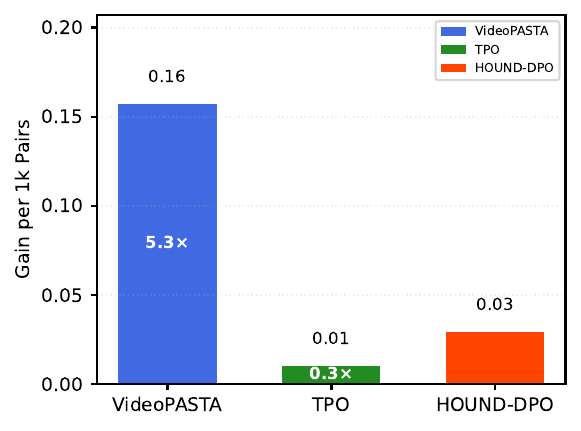}
    \caption{MVBench}
    \label{fig:mvbench_gain}
  \end{subfigure}
  \hfill
  \begin{subfigure}[b]{0.27\linewidth}
    \centering
    \includegraphics[width=\linewidth]{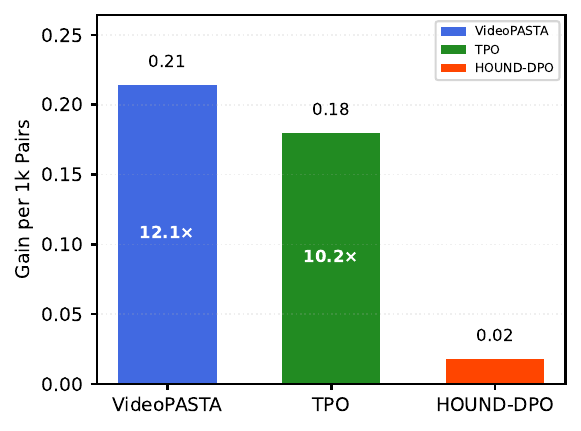}
    \caption{NeXTQA}
    \label{fig:nextqa_gain}
  \end{subfigure}
  \caption{\textbf{Information gain analysis across three representative benchmarks}. Each bar represents performance improvement per 1k preference pairs, calculated as (final score - baseline score) / \# of pairs in thousands.}
  \vspace{-4mm}
  \label{fig:three_benchmarks_gain}
\end{figure*}

%% file: tables/merged_ablation.tex
\begin{table*}[!t]
\centering
\begin{adjustbox}{width=\textwidth,center}
\setlength{\tabcolsep}{1.8pt} 
\renewcommand{\arraystretch}{1.0}
\begin{tabular}{@{}l|ccccccc@{}} 
\toprule
\textbf{Target Model} & \textbf{TempCompass} & \textbf{Perception Test} & \textbf{NeXTQA} & \textbf{MVBench} & \textbf{MLVU} & \textbf{LongVideoBench} & \textbf{VideoMME} \\
\midrule
\rowcolor{gray!15}\multicolumn{8}{c}{(a) \textit{\modelname{} with Self-Generated Preferences}} \\
\midrule
\makecell[l]{Qwen2.5-VL-7B} & 71.2 $\scriptstyle\textcolor{red}{(-0.5)}$ & 68.9 $\scriptstyle\textcolor{darkgreen}{(+0.3)}$ & 76.2 $\scriptstyle\textcolor{darkgreen}{(+0.4)}$ & 65.6 $\scriptstyle\textcolor{darkgreen}{(+0.4)}$ & 68.8 $\scriptstyle\textcolor{darkgreen}{(+0.1)}$ & 60.9 $\scriptstyle\textcolor{darkgreen}{(+0.2)}$ & 63.0 $\scriptstyle\textcolor{darkgreen}{(+0.8)}$ \\
\hline
\makecell[l]{InternVL2.5-8B} & 69.2 $\scriptstyle\textcolor{darkgreen}{(+0.9)}$ & 63.4 $\scriptstyle\textcolor{darkgreen}{(+1.2)}$ & 77.5 $\scriptstyle\textcolor{darkgreen}{(+0.5)}$ & 70.1 $\scriptstyle\textcolor{darkgreen}{(+0.3)}$ & 60.0 $\scriptstyle\textcolor{darkgreen}{(+0.5)}$ & 53.7 $\scriptstyle\textcolor{darkgreen}{(+0.8)}$ & 58.5 $\scriptstyle\textcolor{darkgreen}{(+0.6)}$ \\
\midrule
\rowcolor{gray!15}\multicolumn{8}{c}{\textit{(b) \modelname{} with Preferences from Auxiliary Models}} \\
\midrule

\makecell[l]{Qwen2.5-VL-7B \\  \hspace{3mm}\textcolor{black}{\textit{Query:}} InternVL2.5-38B\\ \hspace{3mm}\textcolor{black}{\textit{Response:}} InternVL2.5-8B} & 71.9 $\scriptstyle\textcolor{darkgreen}{(+0.2)}$ & 69.1 $\scriptstyle\textcolor{darkgreen}{(+0.5)}$ & 76.7 $\scriptstyle\textcolor{darkgreen}{(+0.9)}$ & 65.9 $\scriptstyle\textcolor{darkgreen}{(+0.7)}$ & 69.0 $\scriptstyle\textcolor{darkgreen}{(+0.3)}$ & 61.2 $\scriptstyle\textcolor{darkgreen}{(+0.5)}$ & 63.5 $\scriptstyle\textcolor{darkgreen}{(+1.3)}$ \\
\hline
\makecell[l]{InternVL2.5-8B \\  
\hspace{3mm}\textcolor{black}{\textit{Query:}} InternVL2.5-38B\\ \hspace{3mm}\textcolor{black}{\textit{Response:}} Qwen2.5-VL-7B} & 70.2 $\scriptstyle\textcolor{darkgreen}{(+1.9)}$ & 64.5 $\scriptstyle\textcolor{darkgreen}{(+2.3)}$ & 78.4 $\scriptstyle\textcolor{darkgreen}{(+1.4)}$ & 71.5 $\scriptstyle\textcolor{darkgreen}{(+1.7)}$ & 62.3 $\scriptstyle\textcolor{darkgreen}{(+2.8)}$ & 55.6 $\scriptstyle\textcolor{darkgreen}{(+2.7)}$ & 60.8 $\scriptstyle\textcolor{darkgreen}{(+2.9)}$ \\

\bottomrule
\end{tabular}
\end{adjustbox}
\caption{\textbf{Impact of the source of preference generation.}  (a) The target model generates preference pairs without relying on any external auxiliary models. (b) Two auxiliary models are used to generate query-response (preference) pairs for DPO training the target model. Performance changes ($\scriptstyle\textcolor{darkgreen}{+}$ / $\scriptstyle\textcolor{red}{-}$) are relative to respective DPO target model baselines in Table~\ref{tab:main}.}
\vspace{-1mm}
\label{tab:merged_ablation_studies}
\end{table*}

%% file: tables/ablation_videomme.tex
\begin{table}[!t]
\centering
\begin{adjustbox}{width=\columnwidth,center}
\fontsize{6pt}{7pt}\selectfont 
\setlength{\tabcolsep}{1.2pt} 
\renewcommand{\arraystretch}{1.2} 
\begin{tabular}{@{}lccccc@{}}
\toprule 
\textbf{Method} & \makecell{\textbf{\# of Pref.}\\} & \textbf{Temporal} & \textbf{Spatial} & \textbf{Action} & \textbf{Object} \\
\midrule
\rowcolor{gray!15}\multicolumn{6}{c}{\textit{Baseline Performance}} \\
Qwen2.5-VL & - & 35.0 & 63.6 & 54.0 & 55.9 \\
\midrule
\rowcolor{gray!15}\multicolumn{6}{c}{\textit{Baseline Preference Datasets}} \\
w/ TPO & 10k & \textbf{47.5} $\scriptstyle\textcolor{darkgreen}{(+12.5)}$ & 65.1 $\scriptstyle\textcolor{darkgreen}{(+1.5)}$ & 54.3 $\scriptstyle\textcolor{darkgreen}{(+0.3)}$ & 54.8 $\scriptstyle\textcolor{red}{(-1.1)}$ \\
w/ Hound-DPO & 17k & 42.2 $\scriptstyle\textcolor{darkgreen}{(+7.2)}$ & 61.7 $\scriptstyle\textcolor{red}{(-1.9)}$ & 52.1 $\scriptstyle\textcolor{red}{(-1.9)}$ & 55.5 $\scriptstyle\textcolor{red}{(-0.4)}$ \\
\midrule
\rowcolor{gray!15}\multicolumn{6}{c}{\textit{\modelname{}: Targeting Single Failure Modes}} \\ 
\modelname{} (Temporal Only) & 2.3k & 44.0 $\scriptstyle\textcolor{darkgreen}{(+9.0)}$ & 67.1 $\scriptstyle\textcolor{darkgreen}{(+3.5)}$ & 49.2 $\scriptstyle\textcolor{red}{(-4.8)}$ & 51.3 $\scriptstyle\textcolor{red}{(-4.6)}$ \\
\modelname{} (Spatial Only) & 2.3k & 40.1 $\scriptstyle\textcolor{darkgreen}{(+5.1)}$ & 74.8 $\scriptstyle\textcolor{darkgreen}{(+11.2)}$ & 55.0 $\scriptstyle\textcolor{darkgreen}{(+1.0)}$ & 55.4 $\scriptstyle\textcolor{red}{(-0.5)}$ \\
\modelname{} (Cross-Frame Only) & 2.3k & 43.3 $\scriptstyle\textcolor{darkgreen}{(+8.3)}$ & 66.8 $\scriptstyle\textcolor{darkgreen}{(+3.2)}$ & 54.9 $\scriptstyle\textcolor{darkgreen}{(+0.9)}$ & 57.2 $\scriptstyle\textcolor{darkgreen}{(+1.3)}$ \\
\midrule
\rowcolor{PastaYellow}\textbf{\modelname} \hspace{-0.2em}~\raisebox{-0.2ex} {\includegraphics[height=0.8em]{images/pasta.pdf}} & 
7k & 
45.2 $\scriptstyle\textcolor{darkgreen}{(+10.2)}$ & 
\textbf{78.6} $\scriptstyle\textcolor{darkgreen}{(+15.0)}$ & 
\textbf{56.1} $\scriptstyle\textcolor{darkgreen}{(+2.1)}$ & 
\textbf{58.4} $\scriptstyle\textcolor{darkgreen}{(+2.5)}$ \\
\bottomrule
\end{tabular}
\end{adjustbox}
\caption{\textbf{Effect of targeted failure modes on VideoMME tasks.} Gains/losses relative to baseline Qwen2.5-VL.}
\vspace{-1mm} 
\label{tab:ablation} 
\end{table}

%% file: tables/cross_generalization.tex
\begin{table}[!t]
\setlength{\tabcolsep}{2mm}
\centering
\footnotesize
\begin{tabular}{lcc}
\toprule
\textbf{Model} & \textbf{MovieChat} & \textbf{EgoSchema} \\
\midrule
\rowcolor{gray!15}\multicolumn{3}{c}{\textit{LLaVA-NeXT-Interleave}} \\
Baseline & 40.0 & 51.0 \\
 \rowcolor{PastaYellow}
 + \textbf{\modelname} \hspace{-0.2em}~\raisebox{-0.2ex} {\includegraphics[height=0.8em]{images/pasta.pdf}} & \textbf{41.1} $\scriptstyle\textcolor{darkgreen}{(+1.1)}$ & \textbf{51.9} $\scriptstyle\textcolor{darkgreen}{(+0.9)}$ \\
\midrule
\rowcolor{gray!15}\multicolumn{3}{c}{\textit{LLaVA-OneVision}} \\
Baseline & 44.0 & 64.0 \\
 \rowcolor{PastaYellow}
 + \textbf{\modelname} \hspace{-0.2em}~\raisebox{-0.2ex} {\includegraphics[height=0.8em]{images/pasta.pdf}} & \textbf{45.6} $\scriptstyle\textcolor{darkgreen}{(+1.6)}$ & \textbf{65.0} $\scriptstyle\textcolor{darkgreen}{(+1.0)}$ \\
\midrule
\rowcolor{gray!15}\multicolumn{3}{c}{\textit{InternVL2.5}} \\
Baseline & 46.8 & 52.0 \\
 \rowcolor{PastaYellow}
 + \textbf{\modelname} \hspace{-0.2em}~\raisebox{-0.2ex} {\includegraphics[height=0.8em]{images/pasta.pdf}} & \textbf{47.4} $\scriptstyle\textcolor{darkgreen}{(+0.6)}$ & \textbf{53.6} $\scriptstyle\textcolor{darkgreen}{(+1.6)}$ \\
\midrule
\rowcolor{gray!15}\multicolumn{3}{c}{\textit{Qwen2.5-VL}} \\
Baseline & 44.2 & 57.6 \\
 \rowcolor{PastaYellow}
 + \textbf{\modelname} \hspace{-0.2em}~\raisebox{-0.2ex} {\includegraphics[height=0.8em]{images/pasta.pdf}} & \textbf{46.1} $\scriptstyle\textcolor{darkgreen}{(+1.9)}$ & \textbf{58.1} $\scriptstyle\textcolor{darkgreen}{(+0.5)}$ \\
\bottomrule
\end{tabular}
\caption{\textbf{Cross-Domain Generalization to Unseen Domains.} \modelname{} demonstrates consistent performance gains on movie and egocentric video benchmarks.}
\vspace{-2mm}
\label{tab:cross_domain_generalization}
\end{table}

%% file: tables/mvbench.tex
\begin{table}[!t]
\centering
\begin{adjustbox}{width=\columnwidth,center}
\fontsize{7.5pt}{9pt}\selectfont
\setlength{\tabcolsep}{4pt}
\renewcommand{\arraystretch}{1.2}
\begin{tabular}{@{}lccc@{}}
\toprule
\textbf{MVBench Task} & \textbf{Qwen2.5-VL} & \textbf{Qwen2.5-VL + \modelname{}} & \textbf{Improvement} \\
\midrule
\rowcolor{gray!15}\multicolumn{4}{c}{\textit{Action-Related Tasks}} \\
Action Sequence & 79.5 & \textbf{80.5} & \textcolor{darkgreen}{+1.0} \\
Action Prediction & 67.5 & 67.0 & \textcolor{red}{-0.5} \\
Action Antonym & 87.0 & \textbf{89.5} & \textcolor{darkgreen}{+2.5} \\
Fine-grained Action & 49.5 & \textbf{52.5} & \textcolor{darkgreen}{+3.0} \\
Unexpected Action & 80.5 & \textbf{82.5} & \textcolor{darkgreen}{+2.0} \\
Action Localization & 55.0 & \textbf{67.0} & \textcolor{darkgreen}{+12.0} \\
Action Count & 46.5 & \textbf{53.5} & \textcolor{darkgreen}{+7.0} \\
\midrule
\rowcolor{gray!15}\multicolumn{4}{c}{\textit{Attribute-Related Tasks}} \\
State Change & 58.0 & \textbf{62.0} & \textcolor{darkgreen}{+4.0} \\
Moving Attribute & 90.5 & \textbf{92.0} & \textcolor{darkgreen}{+1.5} \\
\bottomrule
\end{tabular}
\end{adjustbox}
\caption{\textbf{MVBench Sub-task Performance Breakdown.} Improvements on lower-level perceptual tasks support our design rationale of targeting high-level reasoning failures.}
\vspace{-3mm}
\label{tab:mvbench_breakdown}
\end{table}

%% file: sec/X_suppl.tex
\section{DPO Training Dynamics}

\label{dpo_graph}
Figure~\ref{fig:dpo_training} shows the DPO training process for Qwen2.5-VL. The model quickly learns to distinguish between aligned responses ($r^+$) and adversarial ones ($r^-$), as shown by the growing gap between the chosen rewards (\textcolor{darkgreen}{green}, increasing) and the rejected rewards (\textcolor{red}{red}, generally decreasing). At the same time, reward accuracy (\textcolor{blue}{blue}) rises rapidly and stabilizes around 70-75\%, indicating a consistent preference for well-grounded responses over those with targeted misalignments. This demonstrates the effectiveness of our DPO-based alignment approach.

\section{Preference Learning with \modelname{} on Small Models}
\label{scaling_small_models}
\input{tables/abalation_scaling}

To further assess the broad applicability and efficiency of \modelname{}, we evaluate its performance on a range of smaller foundational models, with parameters varying from 1B to 3B. The results, presented in Table~\ref{tab:scaling}, demonstrate that \modelname{} consistently provides performance uplifts even for these more compact architectures. For instance, when applied to Qwen2-VL (2B), \modelname{} improves scores across all seven benchmarks, such as a +1.7 percentage point gain on MVBench (from 60.8 to 62.5) and +1.1 on VideoMME (from 50.1 to 51.2). Similarly, InternVL (1B) + \modelname{} sees gains like +1.3 on MVBench and +0.5 on VideoMME.

These consistent improvements on smaller models highlight several advantages of \modelname{}'s targeted adversarial alignment. Firstly, it highlights that our novel data curation strategy, focusing on specific failure modes (spatial, temporal, cross-frame), provides a learning signal that is effective even for models with lower capacity. Secondly, the ability to boost these smaller models demonstrates that \modelname{} is not solely reliant on the extensive pre-existing knowledge of very large foundation models to achieve its gains but can instill more robust visual reasoning directly. This reinforces the idea that the 7k targeted preference pairs efficiently address core weaknesses, offering a resource-friendly path to enhancing Video-LLMs, making video understanding capabilities more accessible.

\section{Adversarial Robustness}
\label{app:adversarial_robustness}
\input{figures/dpo}
\input{tables/abalation_failure}
To evaluate \modelname{}'s robustness against failure modes, we test 100 videos from LLaVA-Video~\cite{zhang2024video} using GPT-4o~\cite{gpt4o} (prompt provided in Appendix, Figure~
\ref{fig:adversarial-qa-prompt}) to generate both adversarial questions (unanswerable queries) and adversarial options (where ``None of the Above'' is correct) per failure mode. As shown in Table~\ref{tab:adversarial_detailed}, \modelname{} significantly outperforms baselines across all categories, with the most substantial gains in temporal reasoning (+14.5 percentage points). This improved robustness stems directly from our training approach, by exposing the model to targeted adversarial examples during preference optimization, \modelname{} learns to recognize and reject similar misleading inputs during inference. Unlike generic preference optimization, our structured adversarial sampling creates a more discriminative model capable of identifying spatial inconsistencies, temporal contradictions, and cross-frame disconnections. GPT-4o was also used to evaluate model responses (prompt provided in Appendix, Figure~\ref{fig:adversarial-evaluation-prompt}), specifically identifying rejection phrases like ``cannot be answered" and ``insufficient information" when models correctly recognized adversarial inputs.

\section{Qwen2.5-VL Specific Ablations}

\label{app:qwen_specific_ablations}
The following ablations are performed using Qwen2.5-VL as the target model to understand the impact of specific hyperparameter choices within the \modelname{} framework when applied to this backbone:
\subsection{DPO Weight Ratio Analysis}
We explore the impact of different weighting schemes for the DPO loss components (spatial:$\alpha$, temporal:$\beta$, cross-frame:$\gamma$) on Qwen2.5-VL, detailed in Table~\ref{tab:qwen_params_comprehensive}. While focusing on a single dimension (e.g., a 0.6:0.2:0.2 spatial focus) shows some targeted benefits, a more balanced distribution proves superior for overall performance. Our chosen configuration of $\alpha=0.4, \beta=0.4, \gamma=0.2$ (``0.4:0.4:0.2 (Ours)’’) consistently yielded the best results across all seven benchmarks, indicating that while spatial and temporal aspects are crucial, a non-negligible weight for cross-frame reasoning is also important for comprehensive alignment.
\input{tables/qwen_ablation}
\subsection{Number of Adversarial Examples per Aligned Sample}
Table~\ref{tab:qwen_params_comprehensive} shows that performance is optimal with three adversarial examples corresponding to our three targeted failure modes. Using fewer examples leaves certain aspects of video understanding insufficiently challenged, while more examples lead to diminishing returns. This confirms that pairing each aligned response with exactly three adversarial responses, one for each failure mode, best reinforces alignment across spatial, temporal, and cross-frame reasoning.

\subsection{Frame Sampling}
Our analysis of sampling rates (Table~\ref{tab:qwen_params_comprehensive}) shows that using uniformly dense sampling for both aligned and adversarial examples lowers performance as models struggle to detect subtle alignment errors. The optimal configuration (32:1) strikes a balance: dense aligned sampling captures temporal details, while sparse adversarial sampling creates clear misalignment patterns. This result is consistent across benchmarks, highlighting the importance of a well-designed sampling strategy in model training.

\subsection{Image and Video Resolution Settings}
Ablations on image and video resolution for Qwen2.5-VL (Table~\ref{tab:qwen_params_comprehensive}) confirm that higher resolutions generally contribute to better performance, with our selected settings (MAX image resolution $128 \times 28 \times 28$ and VID\_MAX video resolution $64 \times 28 \times 28$) providing a strong balance for our experiments. It is worth noting that score discrepancies with the original Qwen2.5-VL paper may arise because the Qwen team utilized substantially higher input parameters (e.g., video\_max\_pixels up to $768 \times 28 \times 28$, max\_frames up to $768$), which, while potentially beneficial, are often impractical for typical computing environments and our focus on resource-efficient alignment.

\section{Dataset Overview}
\label{app:dataset_overview}
\subsection{Dataset Statistics}
\label{app:dataset_statistics} 
Starting with 3000 videos from ActivityNet~\cite{activitynet}, we systematically generate preference pairs through structured adversarial sampling. For each video \(V\), we generate 10 queries \(Q\) targeting different aspects of video understanding. Each query \(q \in Q\) is paired with three targeted adversarial responses \(r_{\text{spatial}}\), \(r_{\text{temporal}}\), and \(r_{\text{crossframe}}\), representing spatial, temporal, and cross-frame failure modes, respectively.
Theoretically, this setup yields:
\begin{equation}
\label{eq:potential_pairs_appendix_v3} 
\begin{split}
N_{\text{potential}} &= |V| \times |Q| \times |R^-| \\
&= 3000 \times 10 \times 3 = 90,000
\end{split}
\end{equation}
potential preference pairs, where \(|V|\) is the number of videos, \(|Q|\) is the number of queries per video, and \(|R^-|\) is the number of adversarial responses per query.

However, to ensure dataset quality, we employ rigorous filtering using Qwen2.5-32B~\cite{yang2024qwen2} verification using the prompt template given in Figure~\ref{fig:preference-data-filter}. Each preference pair must satisfy three criteria:
\begin{enumerate}
    \item The aligned response should accurately reflect the video content relative to the query.
    \vspace{-0.2cm}
    \item The adversarial response must introduce a clear, deliberate misalignment.
    \vspace{-0.2cm}
    \item The misalignment must be specific to its targeted failure mode.
\end{enumerate}
This verification process retains approximately 7.8\% of the potential pairs (on LLaVA-NeXT-Interleave):
\begin{equation}
\label{eq:final_pairs_appendix_v3}
\begin{aligned}
    N_{\text{final}} &= N_{\text{potential}} \times r_{\text{retention}} \\
                     &\approx 90000 \times 0.078 \approx 7,020.
\end{aligned}
\end{equation}
where \(r_{\text{retention}}\) is the retention rate after quality filtering. This filtered dataset provides a balanced representation across failure modes while maintaining high standards for preference pair quality. The strict filtering ensures that each adversarial example presents a genuine challenge for video-language alignment rather than simple errors or rephrasing.

\subsection{Adversarial Sample Diversity}
\label{app:dataset_samples}
\input{figures/pref_dataset} 
\input{figures/qualitative}
Figure~\ref{fig:supp_examples} illustrates the diversity of adversarial examples demonstrating how \modelname{} targets specific failure modes. These examples were carefully curated to challenge different aspects of video comprehension while maintaining clear distinctions between aligned and adversarial responses.

\paragraph{Spatial Misalignment.} The boat counting example demonstrates our approach to spatial reasoning. While the adversarial response completely negates the presence of obvious visual elements (``no boats''), the challenge lies not in the simple presence/absence but in the precise spatial relationships (``positioned near the shore, with one slightly further out''). This forces the model to develop fine-grained spatial awareness rather than just object detection capabilities.

\paragraph{Temporal Incoherence.} Two examples highlight our approach to temporal understanding. The cooking sequence tests precise transitional timing between steps, where the adversarial response artificially collapses distinct preparation phases into simultaneous actions. Similarly, the equipment preparation example challenges the model's ability to distinguish between sequential and concurrent actions. These adversarial samples are particularly effective because they present plausible but incorrect temporal relationships.

\paragraph{Cross-Frame Disconnection.} The scene transition example illustrates how we assess long-range comprehension. The adversarial response mistakenly interprets superficial visual changes, such as a close-up of a face, as significant narrative shifts, whereas the aligned response accurately identifies meaningful context transitions, like an external threat leading to an internal response. This evaluates the model's ability to track narrative progression across distant frames.

Each example undergoes thorough validation using Qwen2.5-32B~\cite{yang2024qwen2} to ensure that adversarial responses reflect genuine misunderstandings rather than simple errors or rephrasings. This systematic approach to adversarial example generation reinforces robust video-language alignment across multiple dimensions of video understanding.
\section{Qualitative Examples}
\label{app:qualitative_examples}

We present several representative examples that demonstrate how \modelname{} improves video understanding across various scenarios. Figure~\ref{fig:add_qual} illustrates three key aspects of our model's capabilities in handling complex video content.

First, in the camera advertisement sequence, while Qwen2.5-VL~\cite{qwen2.5-VL} fails to recognize the narrative structure and describes it as ``unrelated clips'' \modelname{} successfully captures the purposeful progression from technical camera operation to creative photography. This demonstrates how our cross-frame adversarial sampling helps the model develop a more coherent understanding of extended narratives. Next, the animated sequence with Bugs Bunny showcases \modelname{}'s enhanced ability to track emotional progression. Instead of merely detecting immediate reactions, our model recognizes the escalation from initial irritation to visible anger and, ultimately, to explosive rage. This improvement stems from our temporal incoherence adversarial sampling, which teaches the model to distinguish between simultaneous and sequential emotional states. The cooking demonstration particularly highlights the benefits of our local spatial alignment strategy. While the baseline model confuses the order of preparation steps, \modelname{} correctly identifies the precise sequence of cleaning, coating, and frying the chilies. This accuracy in tracking procedural steps is crucial for practical applications like instructional video understanding. The competition example shows how our model can parse complex sequences of physical and emotional reactions, maintaining temporal coherence even in dynamic scenes. The eclipse footage example reveals \modelname{}'s ability to describe gradual visual transformations accurately, avoiding the baseline's tendency to oversimplify temporal transitions. Finally, the instruction scene identifying magic demonstrates our model's capability to establish clear causal relationships between actions and their outcomes, supported by specific visual evidence.

These qualitative results align with our quantitative findings, showing that \modelname{}'s structured approach to adversarial sampling leads to more precise and accurate video understanding across multiple dimensions. The improvements are especially evident in scenarios requiring temporal coherence, causal reasoning, and the integration of information across extended sequences. The results validate that our adversarial generation approach produces highly targeted examples that specifically challenge the intended aspects of video understanding, creating a focused and efficient learning signal for the model during preference optimization.

\section{Prompt Templates}
\label{app:prompt_templates}
The effectiveness of \modelname{} depends heavily on the careful design of prompts that elicit targeted behaviors from generative models. Our prompt approach focuses on creating a framework that enables the consistent generation of high-quality preference pairs. Rather than using generic prompts that could lead to superficial or inconsistent responses, we develop a hierarchical strategy with explicit constraints and clear objectives. Each template (Figures~\ref{fig:spatial-reasoning-prompt}--\ref{fig:preference-data-filter}) serves a distinct purpose in our pipeline while sharing a common structure that ensures consistency. The spatial misalignment template emphasizes physical relations that remain constant within local temporal windows. The temporal incoherence template focuses on capturing dynamic changes while maintaining causality. The cross-frame disconnection template bridges distant temporal connections without losing local context. Finally, the preference data filtering template acts as a quality control mechanism, ensuring that our generated pairs maintain sufficient contrast while avoiding trivial differences. A key novelty in our method is the explicit incorporation of failure modes into the prompt design itself. Rather than hoping that models will naturally generate useful adversarial examples, we directly encode common pitfalls and misunderstandings into our adversarial prompt variants. The templates are designed to be model-agnostic, allowing them to work with different foundation models while maintaining consistent output quality.

\input{figures/spatial}

\input{figures/temporal}

\input{figures/cross_frame}

\input{figures/filtering}

\input{figures/adversarial_qa}

\input{figures/adversarial_evaluation}

\input{figures/adversarial_question_evaluation}

%% file: tables/abalation_scaling.tex
\begin{table*}[!t]
\centering
\begin{adjustbox}{width=\linewidth,center}
\begin{tabular}{lccccccc}
  \toprule
  \textbf{Model} & \textbf{TempCompass} & \textbf{PerceptionTest} & \textbf{NeXTQA} & \textbf{MVBench} & \textbf{MLVU} & \textbf{LongVideoBench} & \textbf{VideoMME} \\
  \midrule
  Qwen2-VL (2B) & 62.0 & 53.0 & 69.1 & 60.8 & 51.3 & 46.6 & 50.1 \\
\rowcolor{PastaYellow}
  + \textbf{VideoPASTA} \hspace{-0.2em}~\raisebox{-0.2ex} {\includegraphics[height=1em]{images/pasta.pdf}} & 63.6 $\scriptstyle\textcolor{darkgreen}{(+1.6)}$ & 54.4 $\scriptstyle\textcolor{darkgreen}{(+1.4)}$ & 70.6 $\scriptstyle\textcolor{darkgreen}{(+1.5)}$ & 62.5 $\scriptstyle\textcolor{darkgreen}{(+1.7)}$ & 51.9 $\scriptstyle\textcolor{darkgreen}{(+0.6)}$ & 47.7 $\scriptstyle\textcolor{darkgreen}{(+1.1)}$ & 51.2 $\scriptstyle\textcolor{darkgreen}{(+1.1)}$ \\
  \midrule
  Qwen2.5-VL (3B) & 66.6 & 63.1 & 74.9 & 63.5 & 65.9 & 55.8 & 61.2 \\
  \rowcolor{PastaYellow}
  + \textbf{VideoPASTA} \hspace{-0.2em}~\raisebox{-0.2ex} {\includegraphics[height=1em]{images/pasta.pdf}} & 67.3 $\scriptstyle\textcolor{darkgreen}{(+0.7)}$ & 63.9 $\scriptstyle\textcolor{darkgreen}{(+0.8)}$ & 75.6 $\scriptstyle\textcolor{darkgreen}{(+0.7)}$ & 64.5 $\scriptstyle\textcolor{darkgreen}{(+1.0)}$ & 66.7 $\scriptstyle\textcolor{darkgreen}{(+0.8)}$ & 56.0 $\scriptstyle\textcolor{darkgreen}{(+0.2)}$ & 61.8 $\scriptstyle\textcolor{darkgreen}{(+0.6)}$ \\
  \midrule
  InternVL2.5 (1B) & 41.6 & 55.0 & 65.3 & 63.5 & 55.5 & 45.4 & 49.5 \\
  \rowcolor{PastaYellow}
  + \textbf{VideoPASTA} \hspace{-0.2em}~\raisebox{-0.2ex} {\includegraphics[height=1em]{images/pasta.pdf}} & 42.5 $\scriptstyle\textcolor{darkgreen}{(+0.9)}$ & 55.7 $\scriptstyle\textcolor{darkgreen}{(+0.7)}$ & 66.4 $\scriptstyle\textcolor{darkgreen}{(+1.1)}$ & 64.8 $\scriptstyle\textcolor{darkgreen}{(+1.3)}$ & 56.1 $\scriptstyle\textcolor{darkgreen}{(+0.6)}$ & 45.7 $\scriptstyle\textcolor{darkgreen}{(+0.3)}$ & 50.0 $\scriptstyle\textcolor{darkgreen}{(+0.5)}$ \\
  \midrule
  InternVL2.5 (2B) & 47.0 & 57.3 & 68.4 & 65.9 & 56.2 & 48.0 & 53.0 \\
  \rowcolor{PastaYellow}
  + \textbf{VideoPASTA} \hspace{-0.2em}~\raisebox{-0.2ex} {\includegraphics[height=1em]{images/pasta.pdf}} & 48.2 $\scriptstyle\textcolor{darkgreen}{(+1.2)}$ & 58.1 $\scriptstyle\textcolor{darkgreen}{(+0.8)}$ & 69.4 $\scriptstyle\textcolor{darkgreen}{(+1.0)}$ & 67.3 $\scriptstyle\textcolor{darkgreen}{(+1.4)}$ & 56.9 $\scriptstyle\textcolor{darkgreen}{(+0.7)}$ & 48.5 $\scriptstyle\textcolor{darkgreen}{(+0.5)}$ & 54.3 $\scriptstyle\textcolor{darkgreen}{(+1.3)}$ \\
  \bottomrule
\end{tabular}
\end{adjustbox}
    \caption{\textbf{Preference Learning with \modelname{} on small models}.}
    \label{tab:scaling}
\end{table*}

%% file: figures/dpo.tex
\begin{figure}[t]
   \centering
   \includegraphics[width=0.9\columnwidth]{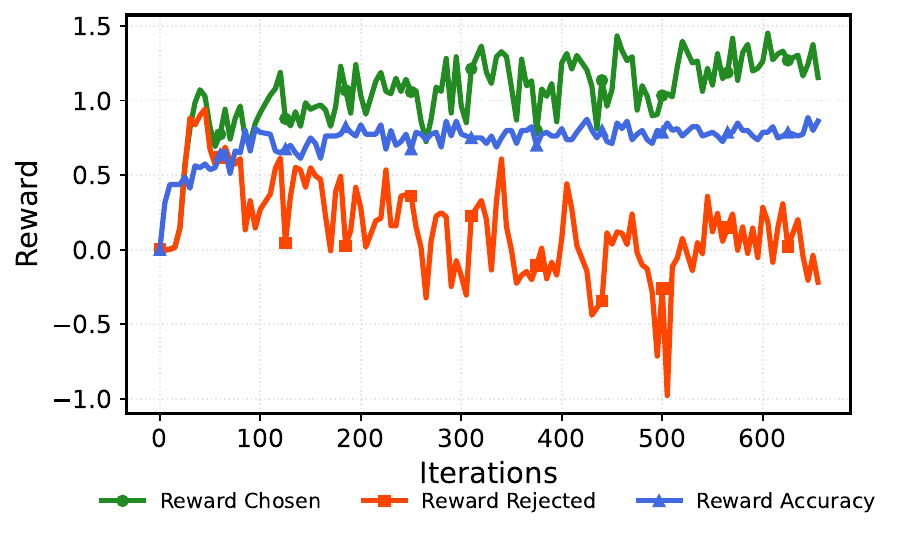}
    \caption{\textbf{DPO training converges on well-grounded responses.}}
   \label{fig:dpo_training}
\end{figure}

%% file: tables/abalation_failure.tex
\begin{table*}[!t]
\centering
\resizebox{\textwidth}{!}{%
\begin{tabular}{lcccccc}
\toprule
\multirow{2}{*}{\textbf{Model}} & \multicolumn{2}{c}{\textbf{Spatial Misalignment}} & \multicolumn{2}{c}{\textbf{Temporal Incoherence}} & \multicolumn{2}{c}{\textbf{Cross-Frame Disconnection}} \\
\cmidrule(lr){2-3} \cmidrule(lr){4-5} \cmidrule(lr){6-7}
& Adv. Question (\%) & Adv. Options (\%) & Adv. Question (\%) & Adv. Options (\%) & Adv. Question (\%) & Adv. Options (\%) \\
\midrule
Qwen2.5-VL~\citep{qwen2.5-VL} & 38.4 & 42.6 & 35.2 & 39.5 & 31.8 & 36.7 \\
LLaVA-Hound-DPO~\citep{zhang2024direct} & 39.2 $\scriptstyle\textcolor{darkgreen}{(+0.8)}$ & 43.1 $\scriptstyle\textcolor{darkgreen}{(+0.5)}$ & 36.5 $\scriptstyle\textcolor{darkgreen}{(+1.3)}$ & 39.8 $\scriptstyle\textcolor{darkgreen}{(+0.3)}$ & 31.9 $\scriptstyle\textcolor{darkgreen}{(+0.1)}$ & 37.2 $\scriptstyle\textcolor{darkgreen}{(+0.5)}$ \\
TPO~\citep{li2025temporal} & 41.3 $\scriptstyle\textcolor{darkgreen}{(+2.9)}$ & 44.5 $\scriptstyle\textcolor{darkgreen}{(+1.9)}$ & 48.2 $\scriptstyle\textcolor{darkgreen}{(+13.0)}$ & 51.4 $\scriptstyle\textcolor{darkgreen}{(+11.9)}$ & 32.4 $\scriptstyle\textcolor{darkgreen}{(+0.6)}$ & 37.5 $\scriptstyle\textcolor{darkgreen}{(+0.8)}$ \\
\midrule
\rowcolor{PastaYellow} \textbf{\modelname} \hspace{-0.2em}~\raisebox{-0.2ex} {\includegraphics[height=1em]{images/pasta.pdf}} & \textbf{46.8} $\scriptstyle\textcolor{darkgreen}{(+8.4)}$ & \textbf{51.1} $\scriptstyle\textcolor{darkgreen}{(+8.5)}$ & \textbf{49.7} $\scriptstyle\textcolor{darkgreen}{(+14.5)}$ & \textbf{52.8} $\scriptstyle\textcolor{darkgreen}{(+13.3)}$ & \textbf{33.1} $\scriptstyle\textcolor{darkgreen}{(+1.3)}$ & \textbf{38.2} $\scriptstyle\textcolor{darkgreen}{(+1.5)}$ \\
\bottomrule
\end{tabular}%
}
\caption{\textbf{Performance on Adversarial QA Samples Across Different Failure Modes}. ``Adv. Question'': Unanswerable queries (higher rejection rate is better). ``Adv. Options'': Questions where ``None of the Above'' is correct (higher NOTA selection is better). Each cell shows correct handling (\%).}

\label{tab:adversarial_detailed}
\end{table*}

%% file: tables/qwen_ablation.tex
\begin{table*}[!t]
\centering
\begin{adjustbox}{width=\textwidth,center}
\setlength{\tabcolsep}{3pt} 
\renewcommand{\arraystretch}{1.2} 
\begin{tabular}{lccccccc}
\toprule
\textbf{Configuration} & \textbf{TempCompass} & \textbf{Perception Test} & \textbf{NeXTQA} & \textbf{MVBench} & \textbf{MLVU} & \textbf{LongVideoBench} & \textbf{VideoMME} \\
\midrule
\rowcolor{gray!15}\multicolumn{8}{c}{\textit{DPO Weight Ratio ($\alpha$:$\beta$:$\gamma$)}} \\
\midrule
0.33:0.33:0.33 (Equal Weights) & 71.5 & 68.2 & 76.8 & 65.7 & 68.5 & 60.6 & 63.4 \\
0.6:0.2:0.2 (Spatial Focus) & 71.6 & 69.3 & 76.5 & 65.5 & 68.1 & 60.3 & 63.1 \\
0.2:0.6:0.2 (Temporal Focus) & 72.2 & 68.1 & 76.3 & 65.4 & 68.7 & 60.8 & 63.2 \\
0.2:0.2:0.6 (Cross-Frame Focus) & 71.2 & 67.8 & 76.9 & 65.8 & 68.9 & 61.2 & 63.5 \\
\textbf{0.4:0.4:0.2 (Ours)} & \textbf{72.3} & \textbf{69.4} & \textbf{77.3} & \textbf{66.3} & \textbf{69.2} & \textbf{61.5} & \textbf{64.1} \\
\midrule
\rowcolor{gray!15}\multicolumn{8}{c}{\textit{Adversarial Examples per Aligned Sample}} \\
\midrule
1 & 71.5 & 68.2 & 76.2 & 65.0 & 67.5 & 58.8 & 62.3 \\
2 & 72.0 & 68.9 & 76.9 & 65.8 & 68.4 & 60.2 & 63.2 \\
\textbf{3 (Ours)} & \textbf{72.3} & \textbf{69.4} & \textbf{77.3} & \textbf{66.3} & \textbf{69.2} & \textbf{61.5} & \textbf{64.1} \\
4 & 72.1 & 69.2 & 77.0 & 66.0 & 68.9 & 61.2 & 63.8 \\
5 & 71.9 & 69.0 & 76.8 & 65.9 & 68.7 & 61.0 & 63.6 \\
\midrule
\rowcolor{gray!15}\multicolumn{8}{c}{\textit{Frame Sampling (Aligned:Adversarial)}} \\
\midrule
32:32 & 71.6 & 68.5 & 76.7 & 65.6 & 68.7 & 60.8 & 62.4 \\
16:16 & 71.5 & 68.4 & 76.6 & 65.5 & 68.6 & 60.7 & 62.3 \\
32:8  & 72.0 & 69.0 & 77.1 & 66.0 & 69.0 & 61.2 & 63.5 \\
16:4  & 71.9 & 68.9 & 77.0 & 65.9 & 68.9 & 61.0 & 63.2 \\
\textbf{32:1 (Ours)} & \textbf{72.3} & \textbf{69.4} & \textbf{77.3} & \textbf{66.3} & \textbf{69.2} & \textbf{61.5} & \textbf{64.1} \\
16:1  & 72.1 & 69.2 & 77.2 & 66.1 & 69.0 & 61.3 & 63.7 \\
\midrule
\rowcolor{gray!15}\multicolumn{8}{c}{\textit{Image Resolution Ablation}} \\
\midrule
MIN=4$\times$28$\times$28, MAX=64$\times$28$\times$28 & 70.1 & 67.2 & 75.2 & 64.1 & 67.3 & 59.4 & 62.0 \\
MIN=4$\times$28$\times$28, MAX=96$\times$28$\times$28 & 71.4 & 68.5 & 76.4 & 65.4 & 68.5 & 60.6 & 63.2 \\
\textbf{MIN=4$\times$28$\times$28, MAX=128$\times$28$\times$28} & \textbf{72.3} & \textbf{69.4} & \textbf{77.3} & \textbf{66.3} & \textbf{69.2} & \textbf{61.5} & \textbf{64.1} \\
\midrule
\rowcolor{gray!15}\multicolumn{8}{c}{\textit{Video Resolution Ablation}} \\
\midrule
VID\_MIN=32$\times$28$\times$28, VID\_MAX=32$\times$28$\times$28 & 70.5 & 67.6 & 75.4 & 64.5 & 67.6 & 59.8 & 62.3 \\
VID\_MIN=48$\times$28$\times$28, VID\_MAX=48$\times$28$\times$28 & 71.6 & 68.7 & 76.5 & 65.6 & 68.6 & 60.8 & 63.5 \\
\textbf{VID\_MIN=64$\times$28$\times$28, VID\_MAX=64$\times$28$\times$28} & \textbf{72.3} & \textbf{69.4} & \textbf{77.3} & \textbf{66.3} & \textbf{69.2} & \textbf{61.5} & \textbf{64.1} \\
\bottomrule
\end{tabular}
\end{adjustbox}
\caption{\textbf{Comprehensive Ablation Studies for VideoPASTA on Qwen2.5-VL.} This table details the impact of DPO weight ratios ($\alpha:\beta:\gamma$), the number of adversarial examples per aligned sample, frame sampling strategies (aligned:adversarial frames), and Qwen specific image/video resolutions. Our chosen configurations are in \textbf{bold}.}
\label{tab:qwen_params_comprehensive}
\end{table*}

%% file: figures/pref_dataset.tex
\begin{figure*}[t!]
    \centering
        \includegraphics[height=18cm, width=15cm]{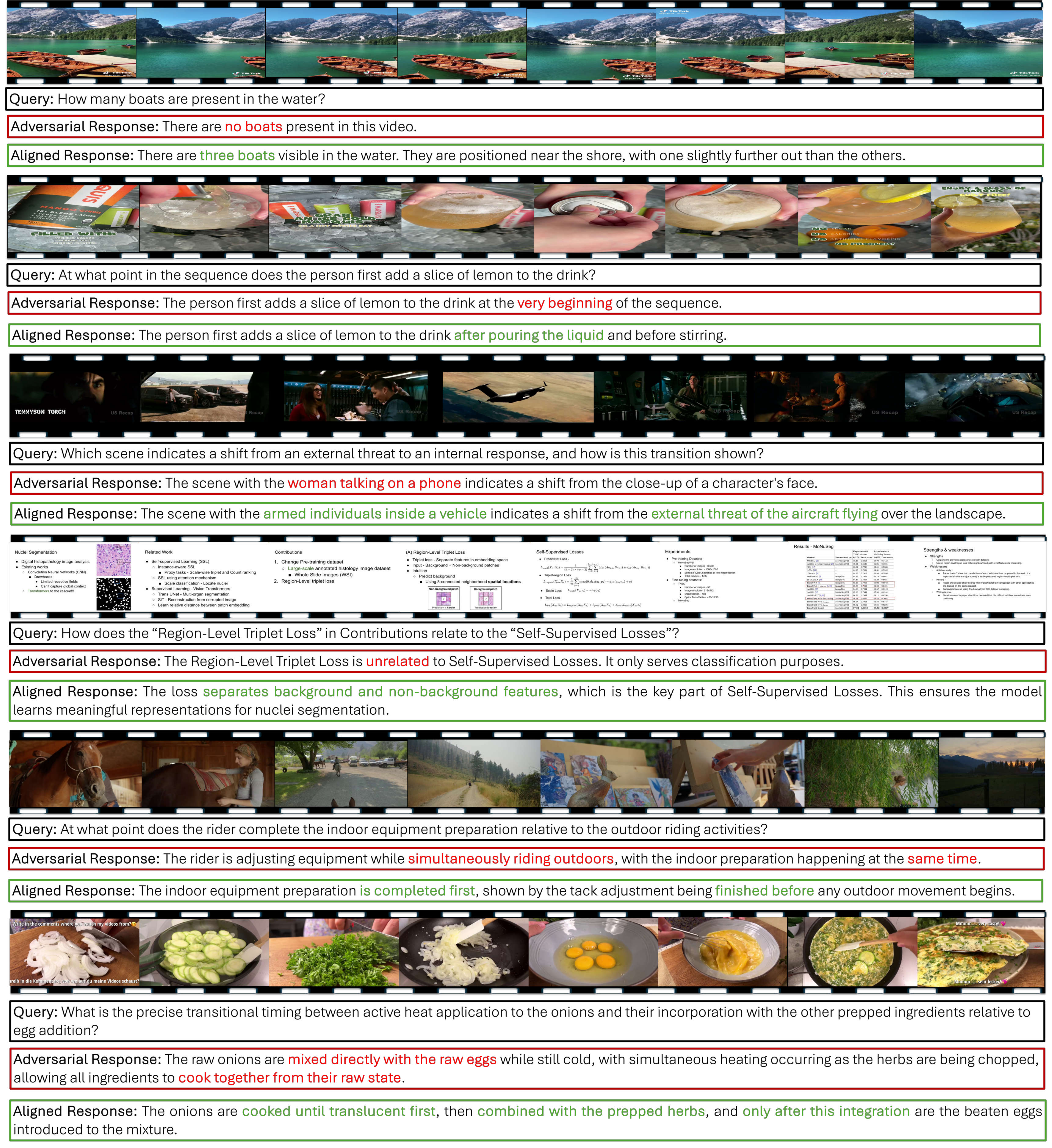}
\caption{\textbf{Adversarial example diversity}. Each row shows a video sequence with its corresponding query and aligned and adversarial responses. The adversarial samples cover \textit{spatial misalignment} (counting objects in scenes), \textit{temporal incoherence} (order of actions in cooking/preparation), and \textit{cross-frame disconnection} (scene transitions and contextual shifts). Adversarial responses deliberately introduce specific misalignments by either negating obvious visual elements, confusing sequential ordering, or collapsing distinct temporal phases into simultaneous events, while aligned responses maintain accurate spatial-temporal alignment with the video content.}
    \label{fig:supp_examples}
\end{figure*}

%% file: figures/qualitative.tex
\begin{figure*}[t!]
    \centering
    \includegraphics[height=18cm, width=15cm]{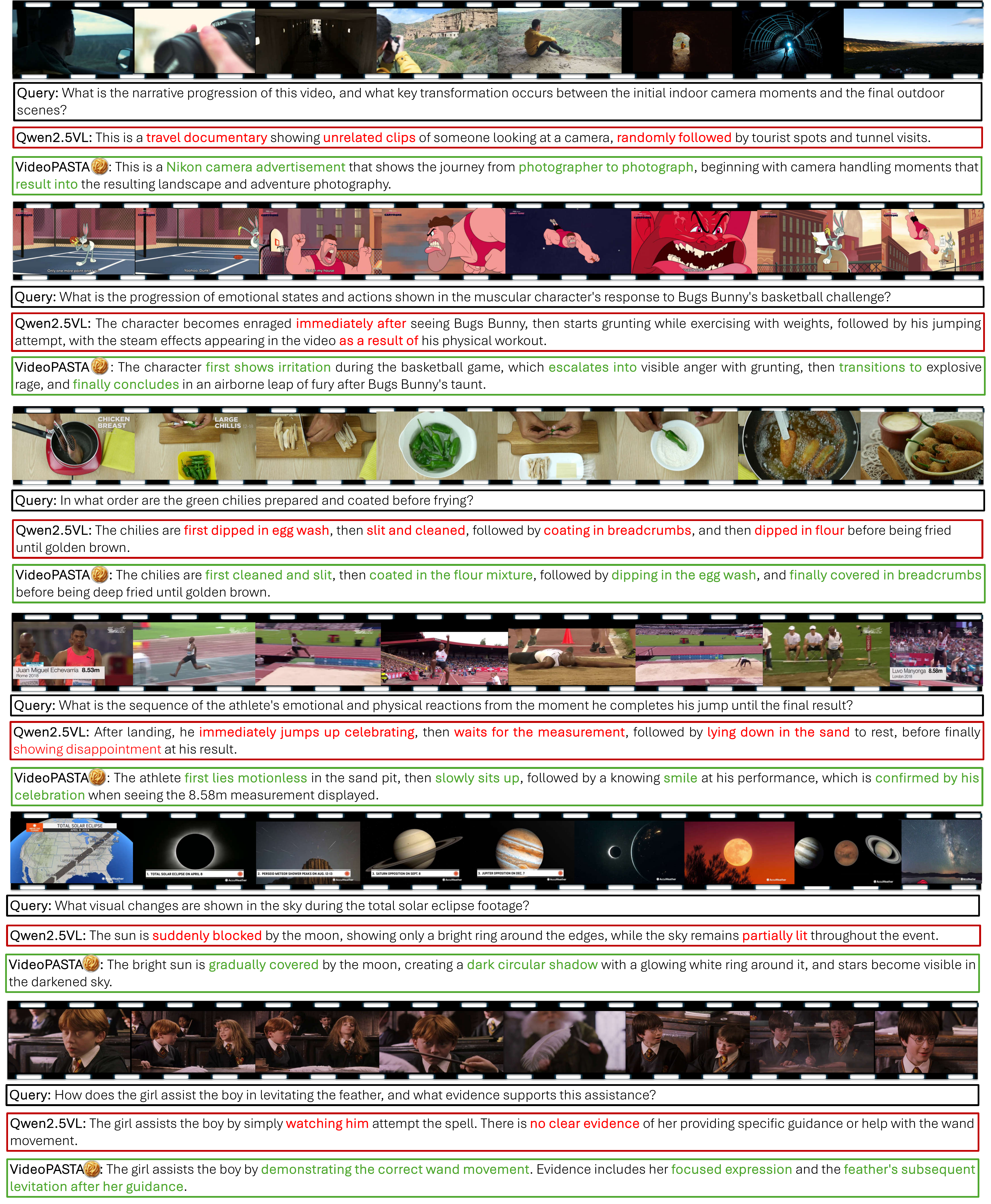}
\caption{\textbf{Qualitative comparison of \modelname{} against Qwen2.5-VL across key failure modes}. The examples demonstrate how our method addresses three critical challenges in video understanding: (1) \textbf{Spatial misalignment} (correctly describing the gradual progression of a solar eclipse and identifying spatial evidence in the Harry Potter scene), (2) \textbf{Temporal incoherence} (accurately capturing sequential emotional progressions in the athlete's reactions and proper cooking preparation steps), and (3) \textbf{Cross-frame disconnection} (maintaining narrative coherence from camera handling to photography outcomes and character emotions). Qwen2.5-VL responses exhibit typical failure patterns: misrepresenting spatial relationships, incorrectly sequencing temporal events, and failing to establish meaningful connections across frames. \modelname{} responses demonstrate robust video-language alignment across all three dimensions.
}
    \label{fig:add_qual}
\end{figure*}

%% file: figures/spatial.tex
\begin{figure*}[htb]
    \centering
    \begin{tcolorbox}[
        title={Spatial Misalignment Prompt},
        colframe=black,
        colback=gray!10,
        coltitle=white,
        fonttitle=\bfseries,
        width=\textwidth
    ]
    You have a single video input. We want to test the model's spatial reasoning according to the following guidelines:
    \begin{enumerate}
        \item \textbf{Aligned Query Generation:}
        \begin{itemize}
            \item Leverage world principles for spatial reasoning to produce 10 queries covering:
            \begin{itemize}
                \item Occlusion (e.g., ``Which object is partially hidden behind another?'')
                \item Depth perception (e.g., ``Which item appears closest to the camera?'')
                \item Relative positioning (``How many objects occupy the left vs.\ right third of the frame?'')
                \item Foreground-background distinctions
                \item Overall frame layout (top vs.\ bottom edges, etc.)
            \end{itemize}
        \end{itemize}
        \item \textbf{Adversarial  Query Generation:}
        \begin{itemize}
            \item For each query, create an adversarial version.
            \item Here, the video will be undersampled at 1 fps.
            \item The adversarial query should actively induce spatial errors.
            \item Example prompts:
            \begin{itemize}
                \item If the query is about occlusion, force the model to claim everything is fully visible
                \item if the query is about depth, insist all objects are equidistant
            \end{itemize}
        \end{itemize}
    \end{enumerate}
    Hence, generate:
    \begin{itemize}
        \item ``Straightforward Spatial Questions'': 10 questions (as if asked under the normal sampling scenario)
        \item ``Adversarial Variants'': 3 matching adversarial instructions (3 per query) that lead the model to produce misaligned/spatially flawed responses.
    \end{itemize}
    \end{tcolorbox}
    \caption{Prompt template for generating aligned and adversarial spatial queries.}
    \label{fig:spatial-reasoning-prompt}
\end{figure*}

%% file: figures/temporal.tex
\begin{figure*}[htb]
    \centering
    \begin{tcolorbox}[
        title={Temporal Incoherence Prompt},
        colframe=black,
        colback=gray!10,
        coltitle=white,
        fonttitle=\bfseries,
        width=\textwidth
    ]
    You have a single video input. We want to test the model's temporal reasoning according to the following guidelines:
    \begin{enumerate}
        \item \textbf{Aligned Query Generation:}
        \begin{itemize}
            \item Leverage world principles for temporal reasoning ability on long videos to produce 10 queries covering:
            \begin{itemize}
                \item Event ordering (e.g., ``Which major action occurs first, and which follows?'')
                \item Action boundaries (e.g., ``Does the person finish one task before starting the next?'')
                \item Transition points (e.g., ``When does the subject switch activities?'')
                \item Causality (e.g., ``Is the second event a direct result of the first?'')
                \item Concurrent actions (e.g., ``Are there any simultaneous events, and how do they overlap?'')
            \end{itemize}
        \end{itemize}
        \item \textbf{Adversarial Query Generation:}
        \begin{itemize}
            \item For each query, create an adversarial version.
            \item Here, the video will be undersampled to induce temporal confusion.
            \item The adversarial query should actively misrepresent event order, action boundaries, or causal links.
            \item Example prompts:
            \begin{itemize}
                \item Claim all actions occur at once, ignoring clear time gaps.
                \item Collapse multiple sequential events into a single continuous action.
            \end{itemize}
        \end{itemize}
    \end{enumerate}
    Hence, generate:
    \begin{itemize}
        \item ``Straightforward Temporal Questions'': 10 questions (as if asked under dense sampling and normal temporal clarity)
        \item ``Adversarial Variants'': 3 matching adversarial instructions (3 per query) that lead the model to produce temporally flawed or misaligned responses.
    \end{itemize}
    \end{tcolorbox}
    \caption{Prompt template for generating aligned and adversarial temporal queries.}
    \label{fig:temporal-reasoning-prompt}
\end{figure*}

%% file: figures/cross_frame.tex
\begin{figure*}[htb]
    \centering
    \begin{tcolorbox}[
        title={Cross-Frame Disconnection Prompt},
        colframe=black,
        colback=gray!10,
        coltitle=white,
        fonttitle=\bfseries,
        width=\textwidth
    ]
    You have a single video input. We want to test the model's cross-frame (long-range) understanding according to the following guidelines:
    \begin{enumerate}
        \item \textbf{Aligned Query Generation:}
        \begin{itemize}
            \item Please produce 10 queries covering:
            \begin{itemize}
                \item Object continuity (e.g., does the same object appear in the opening and closing scenes?)
                \item Character persistence (e.g., which participants return in later segments, and are they consistent with earlier roles?)
                \item Setting evolution (e.g., does the location or environment change over time?)
                \item Repeated actions (e.g., are certain actions performed in distant parts of the video, creating a parallel?)
                \item Foreshadowing (e.g., do early events hint at outcomes shown near the end?)
            \end{itemize}
        \end{itemize}
        \item \textbf{Adversarial Query Generation:}
        \begin{itemize}
            \item For each query, create an adversarial version.
            \item Deliberately break cross-frame connections by forcing the model to ignore continuity or treat identical objects/characters as unrelated.
            \item Example prompts:
            \begin{itemize}
                \item Insist that objects recurring in different scenes are completely different
                \item Claim that characters present at both the start and end have no connection
            \end{itemize}
        \end{itemize}
    \end{enumerate}
    Hence, generate:
    \begin{itemize}
        \item ``Straightforward Cross-Frame Questions'': 10 questions (as if the model respects full continuity across frames)
        \item ``Adversarial Variants'': 3 matching adversarial instructions (3 per query) that lead the model to produce disjointed or inconsistent responses across frames.
    \end{itemize}
    \end{tcolorbox}
    \caption{Prompt template for generating aligned and adversarial queries focusing on cross-frame video understanding.}
    \label{fig:cross-frame-reasoning-prompt}
\end{figure*}

%% file: figures/filtering.tex
\begin{figure*}[htb]
    \centering
    \begin{tcolorbox}[
        title={Preference Data Filtering Prompt},
        colframe=black,
        colback=gray!10,
        coltitle=white,
        fonttitle=\bfseries,
        width=\textwidth
    ]
    You have a single video input and a set of four responses for each query:
    \begin{enumerate}
        \item One \textbf{aligned} response that is claimed to be well-aligned with the video content.
        \item Three \textbf{adversarial} responses, each intentionally introducing spatial, temporal, or cross-frame errors.
    \end{enumerate}

    The goal is to validate that:
    \begin{itemize}
        \item The \emph{aligned} response truly aligns with the query (no unintended contradictions or inaccuracies).
        \item Each \emph{adversarial} response introduces a clear misalignment without merely restating or slightly rephrasing the aligned response.
    \end{itemize}

    For each query and its four responses:
    \begin{enumerate}
        \item \textbf{Sanity-check the aligned response.} 
        \begin{itemize}
            \item Confirm that it accurately reflects the video’s content in relation to the query.
            \item If any errors or contradictions are detected, discard them.
        \end{itemize}

        \item \textbf{Examine each adversarial response.}
        \begin{itemize}
            \item Identify whether it \emph{deliberately} contradicts or distorts the query/video content (e.g., reversed sequence, false spatial claims).
            \item If it is too similar to the aligned response or fails to demonstrate a clear misalignment, discard it.
        \end{itemize}
    \end{enumerate}

    \end{tcolorbox}
    \caption{Prompt template for validating one aligned and three adversarial responses to ensure robust preference pairs.}
    \label{fig:preference-data-filter}
\end{figure*}

%% file: figures/adversarial_qa.tex
\begin{figure*}[htb]
    \centering
    \begin{tcolorbox}[
        title={Adversarial QA Generation Prompt},
        colframe=black,
        colback=gray!10,
        coltitle=white,
        fonttitle=\bfseries,
        width=\textwidth,
        fontupper=\small
    ]
    You are tasked with generating adversarial video question-answering examples to test video language models' robustness. Based on the provided video, create:
    \begin{enumerate}
        \item \textbf{Adversarial Questions:}
        \begin{itemize}
            \item Generate exactly 1 question per failure mode that cannot be reasonably answered from the video content.
            \item These should appear legitimate but contain logical impossibilities or request information that is explicitly not present.
            \item Target the following specific failure modes:
            \begin{enumerate}
                \item \textit{Spatial Misalignment}: Request object relationships that don't exist (e.g., ``How many people are standing behind the blue car?'' when no blue car exists).
                \item \textit{Temporal Incoherence}: Ask about event sequences that violate the timeline (e.g., ``What happens after the person leaves the room?'' when no one leaves).
                \item \textit{Cross-Frame Disconnection}: Request connections between unrelated frames (e.g., ``How does the opening scene connect to the dancing sequence?'' when no dancing occurs).
            \end{enumerate}
        \end{itemize}
        \item \textbf{Adversarial Options:}
        \begin{itemize}
            \item Create exactly 1 multiple-choice question per failure mode where all provided options are incorrect.
            \item Questions should appear legitimate but all options should be misleading.
            \item Include 4 plausible but incorrect options for each question.
            \item The correct answer should always be ``None of the Above'' (not included in the options).
            \item Target the same three failure modes as above.
        \end{itemize}
    \end{enumerate}
    Format each output as:
    \begin{itemize}
        \item \textbf{Adversarial Question [Spatial Misalignment]:} [Question text].
        \item \textbf{Adversarial Question [Temporal Incoherence]:} [Question text].
        \item \textbf{Adversarial Question [Cross-Frame Disconnection]:} [Question text].
        \item \textbf{Adversarial Options [Spatial Misalignment]:}\\
        Question: [Question text].\\
        Options:\\
        A. [Option 1].\\
        B. [Option 2].\\
        C. [Option 3].\\
        D. [Option 4].\\
        Correct Answer: None of the Above.
        \item \textbf{Adversarial Options [Temporal Incoherence]:}\\
        Question: [Question text].\\
        Options:\\
        A. [Option 1].\\
        B. [Option 2].\\
        C. [Option 3].\\
        D. [Option 4].\\
        Correct Answer: None of the Above.
        \item \textbf{Adversarial Options [Cross-Frame Disconnection]:}\\
        Question: [Question text].\\
        Options:\\
        A. [Option 1].\\
        B. [Option 2].\\
        C. [Option 3].\\
        D. [Option 4].\\
        Correct Answer: None of the Above.
    \end{itemize}
    \end{tcolorbox}
    \caption{Prompt template for generating adversarial questions and options across different failure modes.}
    \label{fig:adversarial-qa-prompt}
\end{figure*}

%% file: figures/adversarial_evaluation.tex
\begin{figure*}[htb]
    \centering
    \begin{tcolorbox}[
        title={Adversarial Question Evaluation Prompt},
        colframe=black,
        colback=gray!10,
        coltitle=white,
        fonttitle=\bfseries,
        width=\textwidth
    ]
    You are an expert evaluator for video question-answering systems. Your task is to determine whether a model correctly identifies unanswerable questions about a video.
    
    \textbf{Input:}
    \begin{enumerate}
        \item Video context (description provided).
        \item Adversarial question (deliberately unanswerable).
        \item Model response to the question.
    \end{enumerate}
    
    \textbf{Evaluation criteria:}
    \begin{enumerate}
        \item A correct response must identify that the question cannot be answered from the video.
        \item Look for specific rejection patterns:
        \begin{enumerate}
            \item Explicit rejection phrases: ``cannot be answered,'' ``not possible to determine,'' ``not shown in the video.''
            \item Identification of missing elements: ``no such [object/action] appears,'' ``insufficient information.''
            \item Recognition of logical impossibility: ``the premise of the question is incorrect,'' ``this scenario does not occur.''
        \end{enumerate}
        \item Incorrect responses attempt to answer the question by:
        \begin{enumerate}
            \item Hallucinating non-existent objects, actions, or relationships.
            \item Making definitive statements about information not present in the video.
            \item Failing to identify the adversarial nature of the question.
        \end{enumerate}
    \end{enumerate}
        \textbf{Output format:}
    \begin{enumerate}
        \item \textbf{Judgment}: [CORRECT/INCORRECT].
        \item \textbf{Reasoning}: Brief justification for your evaluation (1-2 sentences).
        \item \textbf{Rejection Keywords Identified}: List specific rejection phrases used by the model.
    \end{enumerate}
    
    Provide a binary decision (CORRECT/INCORRECT) based strictly on whether the model appropriately identified the question as unanswerable.
    \end{tcolorbox}
    \caption{Prompt template for evaluating model responses to adversarial questions.}
    \label{fig:adversarial-evaluation-prompt}
\end{figure*}

%% file: figures/adversarial_question_evaluation.tex
\begin{figure*}[t]
    \centering
    \begin{tcolorbox}[
        title={Adversarial Example Evaluation Prompt},
        colframe=black,
        colback=gray!10,
        coltitle=white,
        fonttitle=\bfseries,
        width=\textwidth
    ]
    \textbf{Task}: Evaluate whether the provided adversarial example correctly targets its intended failure mode in video understanding.
        
    \textbf{Query}: [Original question asked about the video]
    
    \textbf{Aligned Response}: [The correct/preferred response to the query]
    
    \textbf{Adversarial Example}: [The adversarial example to be evaluated]
    
    \textbf{Claimed Failure Mode}: [One of: ``Spatial Misalignment'', ``Temporal Incoherence'', or ``Cross-Frame Disconnection'']
    
    \textbf{Failure Mode Definitions}:
    \begin{itemize}
        \item \textbf{Spatial Misalignment}: Incorrectly describing spatial relations, object positions, occlusion patterns, depth, or relative positioning within a single frame.
        \item \textbf{Temporal Incoherence}: Violating the natural ordering of events, describing sequential actions as simultaneous, merging distinct events, or misordering the sequence of activities shown in the video.
        \item \textbf{Cross-Frame Disconnection}: Breaking object persistence across frames, describing the same object as different entities across scenes, failing to maintain character/object consistency, or incorrectly describing changes between distant frames.
    \end{itemize}
    
    \textbf{Evaluation Instructions}:
    \begin{enumerate}
        \item Carefully analyze the adversarial example in relation to the aligned response.
        \item Determine if the adversarial example genuinely induces the claimed failure mode.
        \item Your evaluation should be based solely on the definitions provided above.
        \item Provide a binary judgment: ``Yes'' if the adversarial example correctly targets the claimed failure mode, ``No'' if it does not.
        \item Briefly explain your reasoning (2-3 sentences).
    \end{enumerate}
    
    \textbf{Output Format}:
    \begin{verbatim}
    Judgment: [Yes/No]
    Reasoning: [Your brief explanation]
    \end{verbatim}
    \end{tcolorbox}
    \caption{Prompt used for evaluating whether adversarial examples correctly target their claimed failure modes.}
    \label{fig:evaluation-prompt}
\end{figure*}